\documentclass[sigconf]{acmart}

\usepackage{booktabs} 
\usepackage[utf8]{inputenc} 
\usepackage[T1]{fontenc}    
\usepackage{cleveref}       
\usepackage{url}            
\usepackage{booktabs}       
\usepackage{amsfonts}       
\usepackage{nicefrac}       
\usepackage{microtype}      
\usepackage{amsmath}
\usepackage{bm}
\usepackage{graphicx}
\usepackage{subfigure}
\usepackage[english]{babel}
\usepackage{blindtext}
\usepackage{multicol}
\usepackage[group-separator={,}]{siunitx}
\usepackage[ugly]{units}
\usepackage{array}
\usepackage{hyperref}
\newcolumntype{C}{>{\lower.2ex\hbox to 2ex\bgroup\hss}c<{\hss\egroup}}

\newcommand{\x}{\mathbf{x}}
\newcommand{\y}{\mathbf{y}}

\newcommand{\T}{\mathcal{T}}
\newcommand{\W}{\mathcal{W}}
\newcommand{\X}{\mathcal{X}}
\newcommand{\Y}{\mathcal{Y}}

\newcommand{\triplet}{(i,j,k)}

\newcommand{\pt}{p^{\scriptscriptstyle (t')}_{\scriptscriptstyle ijk}}
\newcommand{\method}{$t$-ETE}

\newcommand{\veps}{\ell}
\newcommand{\loss}{\ell_{\scriptscriptstyle ijk}{\scriptscriptstyle (\Y)}}
\newcommand{\tloss}{\ell^{\scriptscriptstyle (t^\prime)}_{\scriptscriptstyle ijk}{\scriptscriptstyle (\Y)}}

\newcommand{\Loss}{L_{\scriptscriptstyle \T}}
\newcommand{\cLoss}{C_{\scriptscriptstyle \T}}
\newcommand{\cLossw}{C_{\scriptscriptstyle \W,\T}}

\usepackage{natbib}
\usepackage{algorithm}
\usepackage{algorithmic}

\begin{document}
\title{Low-dimensional Data Embedding via Robust Ranking}

\author{Ehsan Amid}
\affiliation{%
  \institution{University of California,}
  \city{Santa Cruz} 
  \state{CA} 
  \postcode{95064}
}
\email{eamid@ucsc.edu}

\author{Nikos Vlassis}
\affiliation{%
  \institution{Adobe Research}
  \city{San Jose} 
  \state{CA} 
  \postcode{95113}
}
\email{vlassis@adobe.com}

\author{Manfred K. Warmuth}
\affiliation{%
\institution{University of California,}
  \city{Santa Cruz} 
  \state{CA} 
  \postcode{95064}
  }
\email{manfred@ucsc.edu}


\begin{abstract}

We describe a new method called \method\
for finding a low-dimensional embedding 
of a set of objects in Euclidean space. 
We formulate the embedding problem as a joint ranking problem over a set of triplets, where each triplet captures the relative similarities between three objects in the set.
By exploiting recent advances in robust ranking, \method\
produces high-quality embeddings even in the presence of
a significant amount of noise and better preserves local
scale than known methods, such as t-STE and t-SNE. 
In particular, our method produces significantly better results than t-SNE 
on signature datasets while also being faster to compute.
\end{abstract}




\maketitle

\section{Introduction}

Learning a metric embedding for a set of objects based on
relative similarities is a central problem in human
computation and crowdsourcing. The application domain
includes a variety of different fields such as recommender
systems and psychological questionnaires. The relative
similarities are usually provided in the form of 
triplets, where a triplet $\triplet$ expresses that
``\emph{object~$i$ is more similar to object~$j$ than to
    object~$k$}'', for which the similarity function may be
unknown or not even quantified. The first object $i$ is
referred to as the \emph{query} object and objects
$j$ and $k$ are the \emph{test} objects. The
triplets are typically gathered by human evaluators 
via a data-collecting mechanism such as Amazon Mechanical Turk\footnote{\url{https://www.mturk.com}}. These types of constraints have also been used as side information in semi-supervised metric learning~\cite{itml,lsml} and clustering~\cite{sklr}. 

Given a set of relative similarity comparisons on a set of objects, the goal of triplet embedding is to find a representation for the objects in some metric space such that the constraints induced by the triplets are satisfied as much as possible. In other words, the embedding should reflect the underlying similarity function from which the constraints were generated. Earlier methods for triplet embedding include Generalized Non-metric Multidimensional Scaling (GNMDS)~\cite{gnmds}, Crowd Kernel Learning (CKL)~\cite{ckl}, and Stochastic Triplet Embedding (STE) and extension, t-distributed STE (t-STE)~\cite{ste}. 

\begin{figure*}
\begin{center}
\centering\sf\setlength{\tabcolsep}{59pt}
\begin{tabular}{|*{4}{C|}}
\hline
  \subfigure[]{\includegraphics[width=0.245\textwidth]{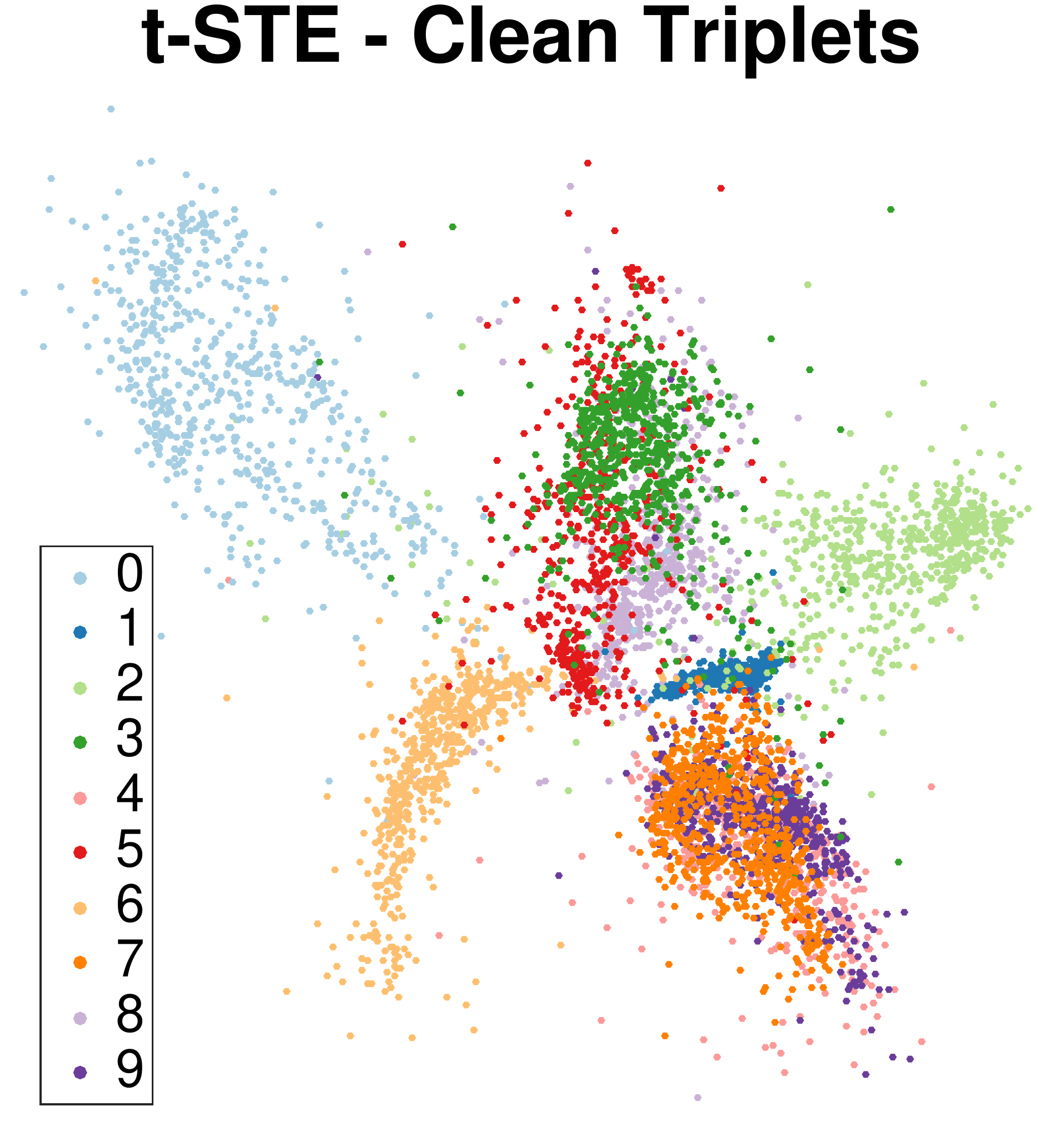}\label{fig:mnist-tste-clean}} &   \subfigure[]{\includegraphics[width=0.245\textwidth]{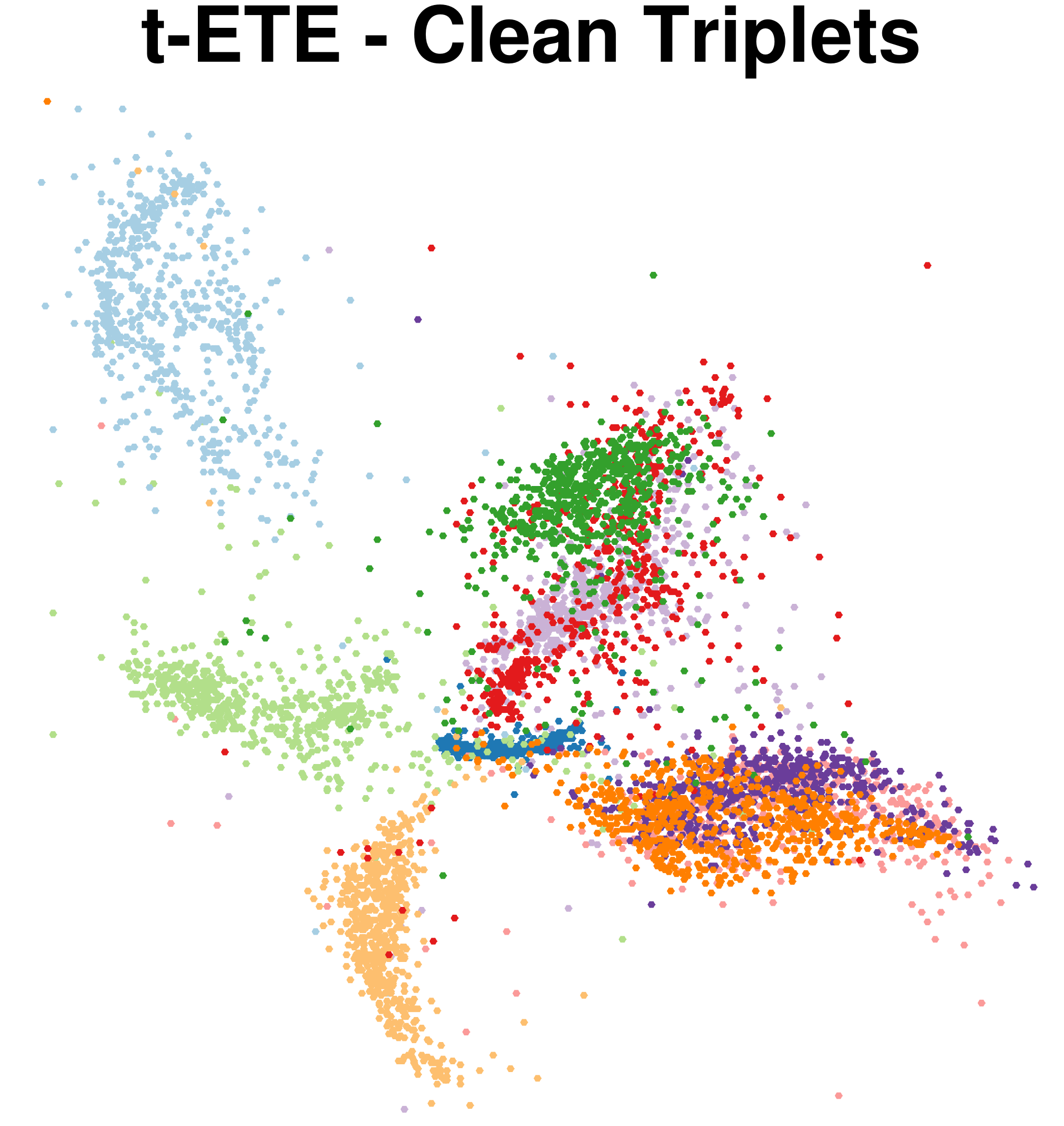}\label{fig:mnist-tete-clean}} & \subfigure[]{\includegraphics[width=0.245\textwidth]{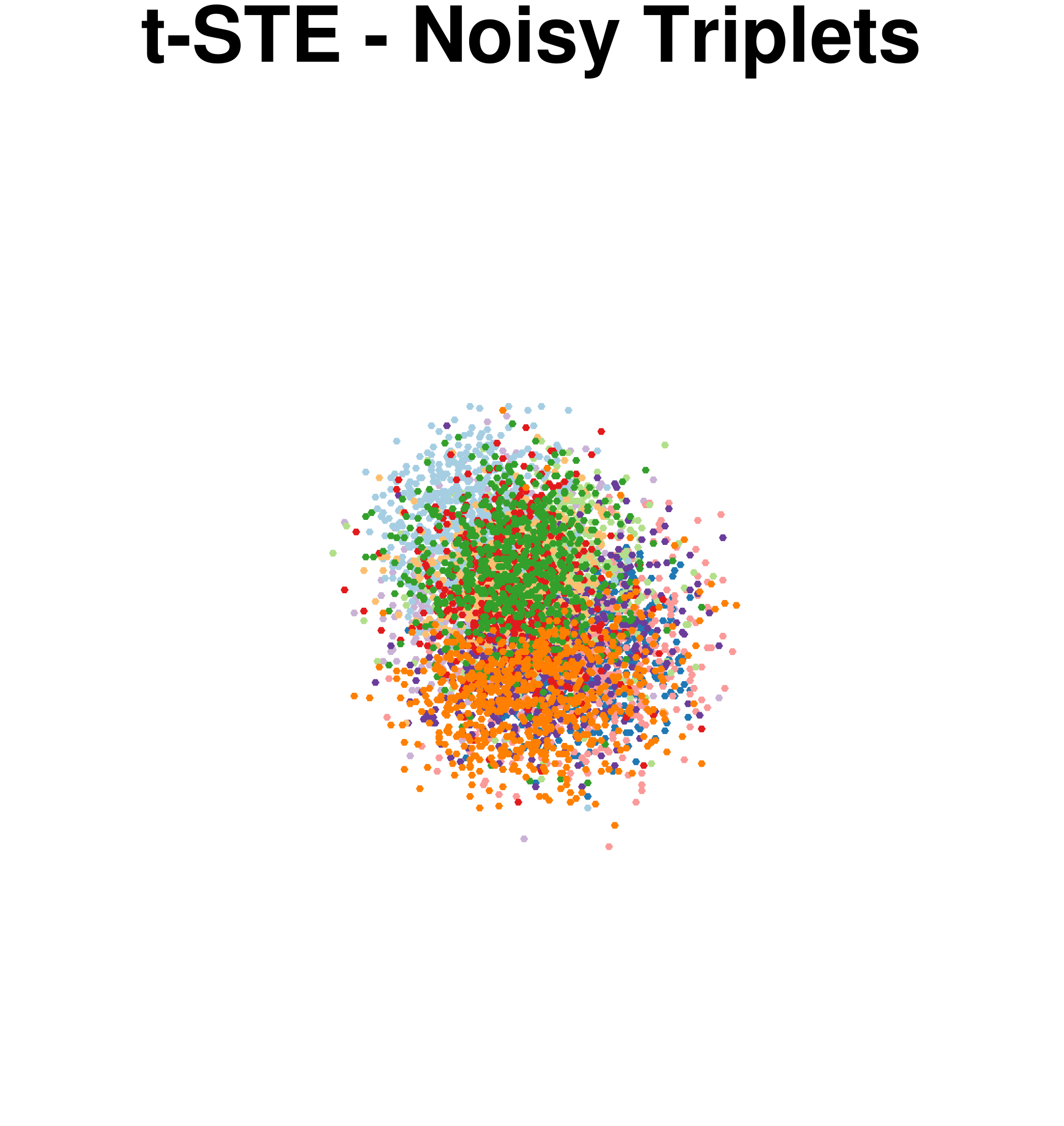}\label{fig:mnist-tste-noise}} & \subfigure[]{\includegraphics[width=0.245\textwidth]{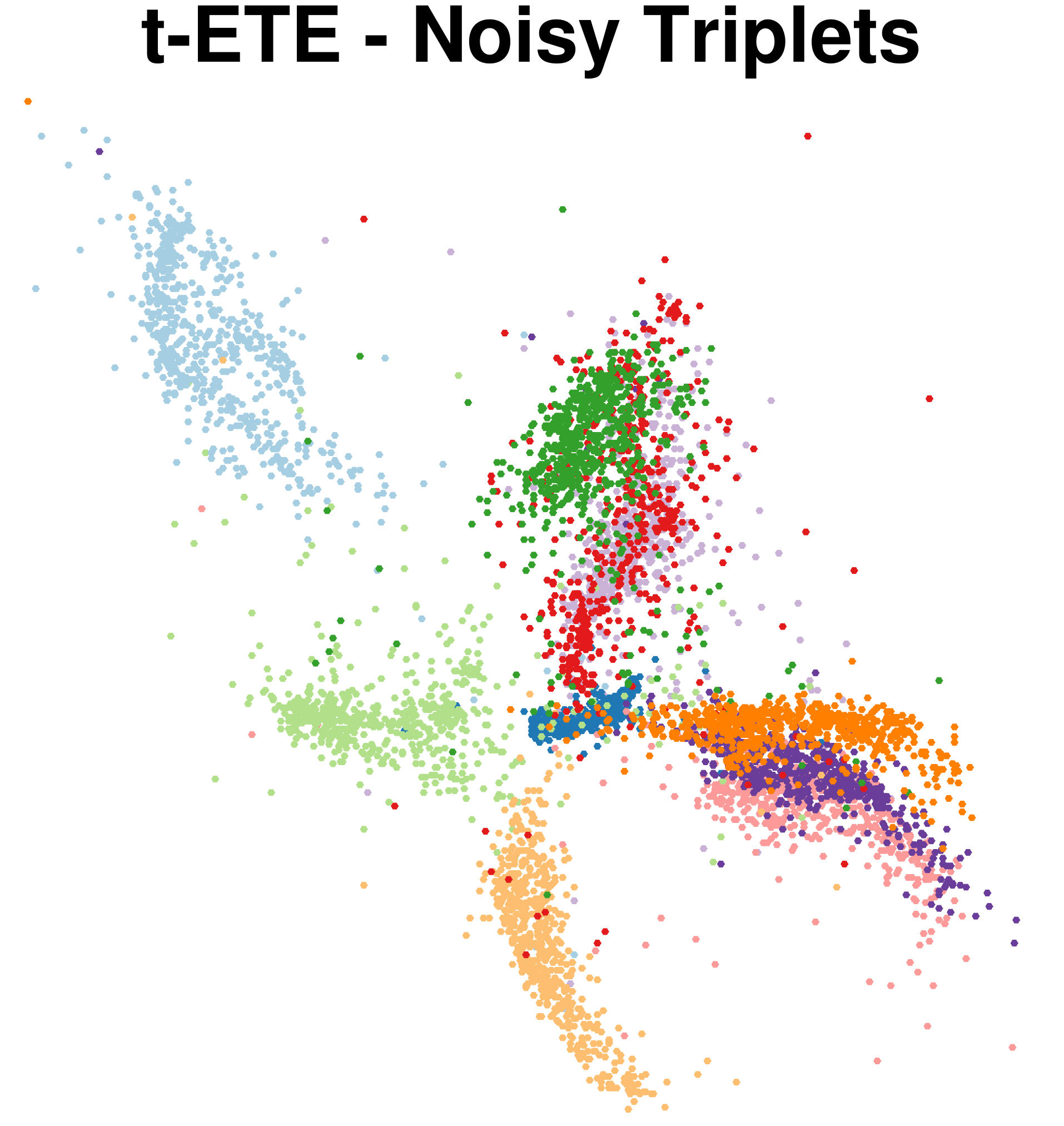}\label{fig:mnist-tete-noise}}\\\hline
\end{tabular}
\caption{Experiments on the MNIST dataset: noise-free triplets using (a) t-STE, and (b) $t$-ETE, and triplets with $20\%$ noise using (c) t-STE, and (d) the proposed $t$-ETE.}\label{fig:noise-mnist}
\end{center}
\end{figure*}

One major drawback of the previous methods for triplet
embedding is that their performance can drop significantly
when a small amount of noise is introduced in the data. The
noise may arise due to different reasons. For instance,
each human evaluator may use a different similarity function 
when comparing objects~\cite{mvte}. 
As a result, there might exist conflicting triplets 
with reversed test objects. 
Another type of noise could be due to the insufficient degree of freedom 
when mapping an intrinsically (and possibly hidden) high-dimensional representation to a lower-dimensional embedding. A simple example is mapping uniformly distributed points on a two-dimensional circle to a one-dimensional line; regardless of the embedding, the end points of line will always violate some similarity constraints. 

In this paper, we cast the triplet embedding problem as a joint ranking problem.
In any embedding, for each object $i$, the remaining object are naturally
ranked by their ``distance'' to $i$. The triplet $\triplet$
expresses that the object $j$ should be ranked higher than object $k$ 
for the ranking of $i$. Therefore, triplet embedding 
can be viewed as mapping the objects into a Euclidean space
so that the joint rankings belonging to all query
objects are as consistent (with respect to the triplets) as possible. 
In order to find the embedding, we define a loss for each triplet
and minimize the sum of losses over all triplets. 
Initially our triplet loss is unbounded.
However in order to make our method robust to noise, 
we apply a novel robust transformation (using the
generalized $\log$ function), which caps the triplet loss by a constant. 
Our new method, $t$-Exponential Triplet
Embedding ($t$-ETE)\footnote{The acronym t-STE is based on the Student-t distribution. Here, ``t'' is part of the name of the distribution. Our method, $t$-ETE, is based on $t$-exponential family. Here, $t$ is a parameter of the model.}, inherits the heavy-tail properties of t-STE in producing high-quality embeddings, while being significantly more robust to noise than any other method. Figure~\ref{fig:noise-mnist} illustrates examples of embeddings of a subset of $6000$ data points from the MNIST dataset using t-STE and our proposed method. The triplets are synthetically generated by sampling a random point from one of the $20$-nearest neighbors for each point and another point from those that are located far away ($100$ triplets for each point). The two embeddings are very similar when there is no noise in the triplets (Figures~\ref{fig:mnist-tste-clean} and \ref{fig:mnist-tete-clean}). However, after `reversing' $20\%$ of the triplets, t-STE fails to produce a meaningful embedding (Figure~\ref{fig:mnist-tste-noise}) while $t$-ETE is almost unaffected by the noise (Figure~\ref{fig:mnist-tete-noise}).

We also apply our \method\ method to dimensionality reduction and develop a new technique, which samples a subset of triplets in the high-dimensional space and finds the low-dimensional representation that satisfies the corresponding ranking. We quantify the importance of each triplet by a non-negative weight. We show that even a small carefully chosen subset of triplets capture sufficient information about the local as well as the global structure of the data to produce high-quality embeddings. Our proposed method outperforms the commonly used t-SNE~\cite{tsne} for dimensionality reduction in many cases while having a much lower complexity.

\section{Triplet Embedding via Ranking} \label{sec:trip}

In this section we formally define the triplet embedding problem. Let $\mathcal{I} = \{1,2,\ldots, N\}$ denote a set of objects.
Suppose that the feature (metric) representation of these objects is unknown. However, some information about the relative similarities of these objects is available in the form of \emph{triplets}.
A triplet $(i,j, k)$ is an ordered tuple which represents a constraint on the relative similarities of the objects $i$, $j$, and $k$, of the type
``object $i$ is more similar to object $j$ than to object~$k$.''
Let $\T = \{\triplet\}$ denote the set of triplets available for the set of objects $\mathcal{I}$.

Given the set of triplets $\T$, the \emph{triplet embedding} problem amounts to finding a metric representation of the objects, $\Y = \{\y_1,\y_2,\ldots,\y_N\}$, such that the similarity constraints imposed by the triplets are satisfied as much as possible by a given distance function in the embedding. For instance, in the case of Euclidean distance, we want
\begin{equation}
\triplet \Longrightarrow \Vert \y_i - \y_j \Vert < \Vert \y_i - \y_k \Vert\,,  \text{ w.h.p.}
\end{equation}
The reason that we may not require all the constraints to be satisfied in the embedding is that there may exist inconsistent and/or conflicting constraints among the set of triplets.
This is a very common phenomenon when the triplets are collected via human evaluators via crowdsourcing~\cite{food,mvte}.

We can consider the triplet embedding problem as a ranking problem imposed by the set of constraints $\T$. More specifically, each triplet $\triplet$ can be seen as a partial ranking result where for a query over $i$, we are given two results, namely $j$ and $k$, and the triplet constraint specifies that \emph{``the result $j$ should have relatively higher rank than $k$''}. In this setting, only the order of closeness of test objects to the query object determines the ranking of the objects.

Let us define $\loss \in [0, \infty)$ to be non-negative loss associated with the triplet constraint $\triplet$. To reflect the ranking constraint, the loss $\loss$ should be a monotonically increasing (decreasing) function of the pairwise distance $\Vert \y_i - \y_j \Vert$ ($\Vert \y_i - \y_k \Vert$). These properties ensure that $\loss \rightarrow 0$  whenever $\Vert \y_i - \y_j \Vert \rightarrow 0$ and $\Vert \y_i - \y_k \Vert \rightarrow \infty$. We can now define the triplet embedding problem as minimizing the sum of the ranking losses of the triplets in $\T$, that is,
\begin{equation}
\label{eq:obj-naive}
\min_{\Y}\, \Loss,\quad \Loss = \sum_{\triplet \in \T} \loss\, .
\end{equation}
In the above formulation, the individual loss of each triplet is unbounded. This means that in cases where a subset of the constraints are corrupted by noise, the loss of even a single inconsistent triplet may dominate the total objective~\eqref{eq:obj-naive} and result in a poor performance.  In order to avoid such effect, we introduce a new robust transformation to cap the individual loss of each triplet from above by a constant. As we will see, the capping helps to avoid the noisy triplets and produce high-quality embeddings, even in the presence of a significant amount of noise.

\section{Robust Loss Transformations}
\label{sec:trans}

\begin{figure}[t!]	
	\subfigure[]{\includegraphics[width=0.22\textwidth]{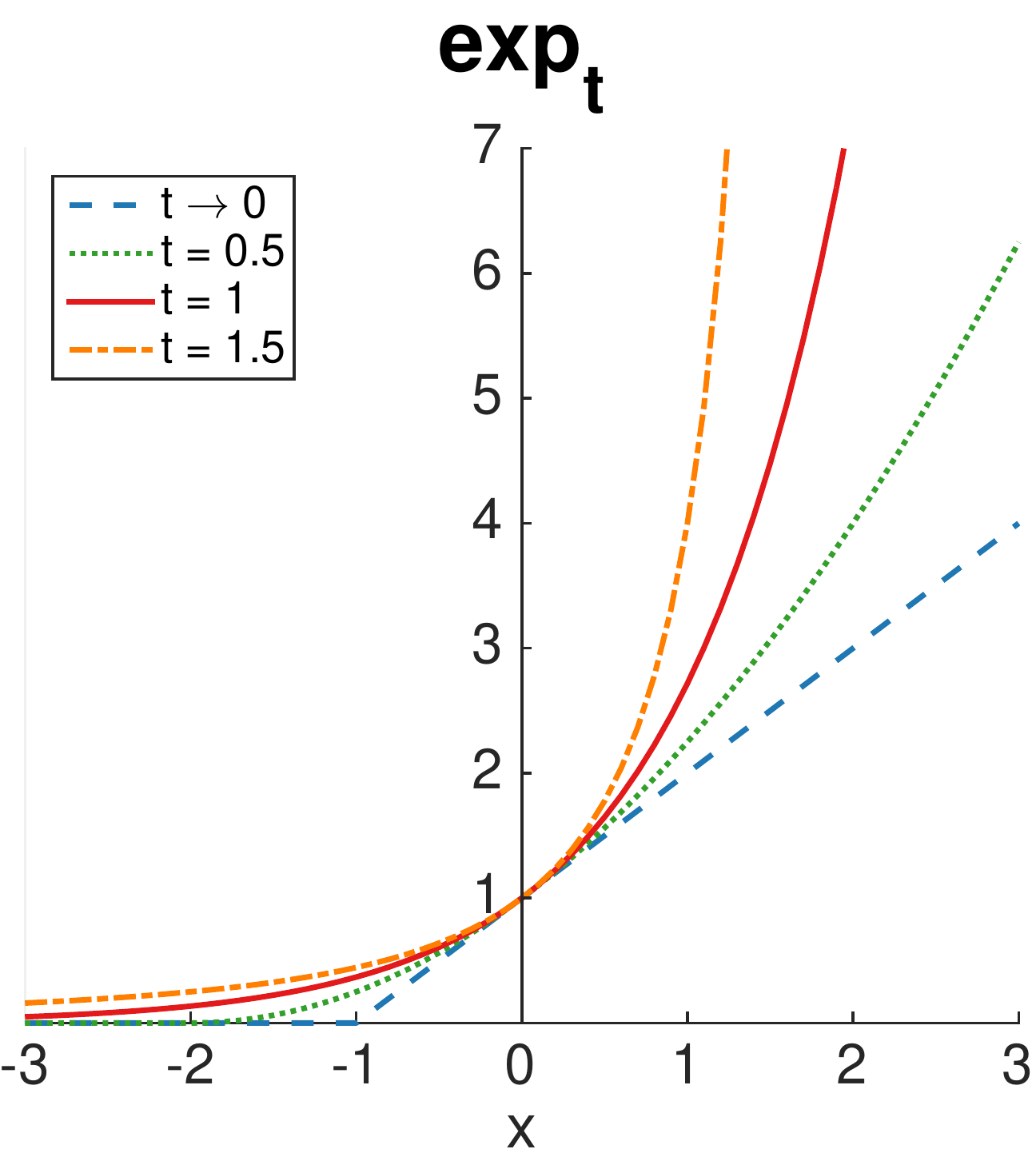}\label{fig:expt}}\hspace{5mm}
     \subfigure[]{\raisebox{0mm}[0pt][0pt]{\includegraphics[width=0.22\textwidth]{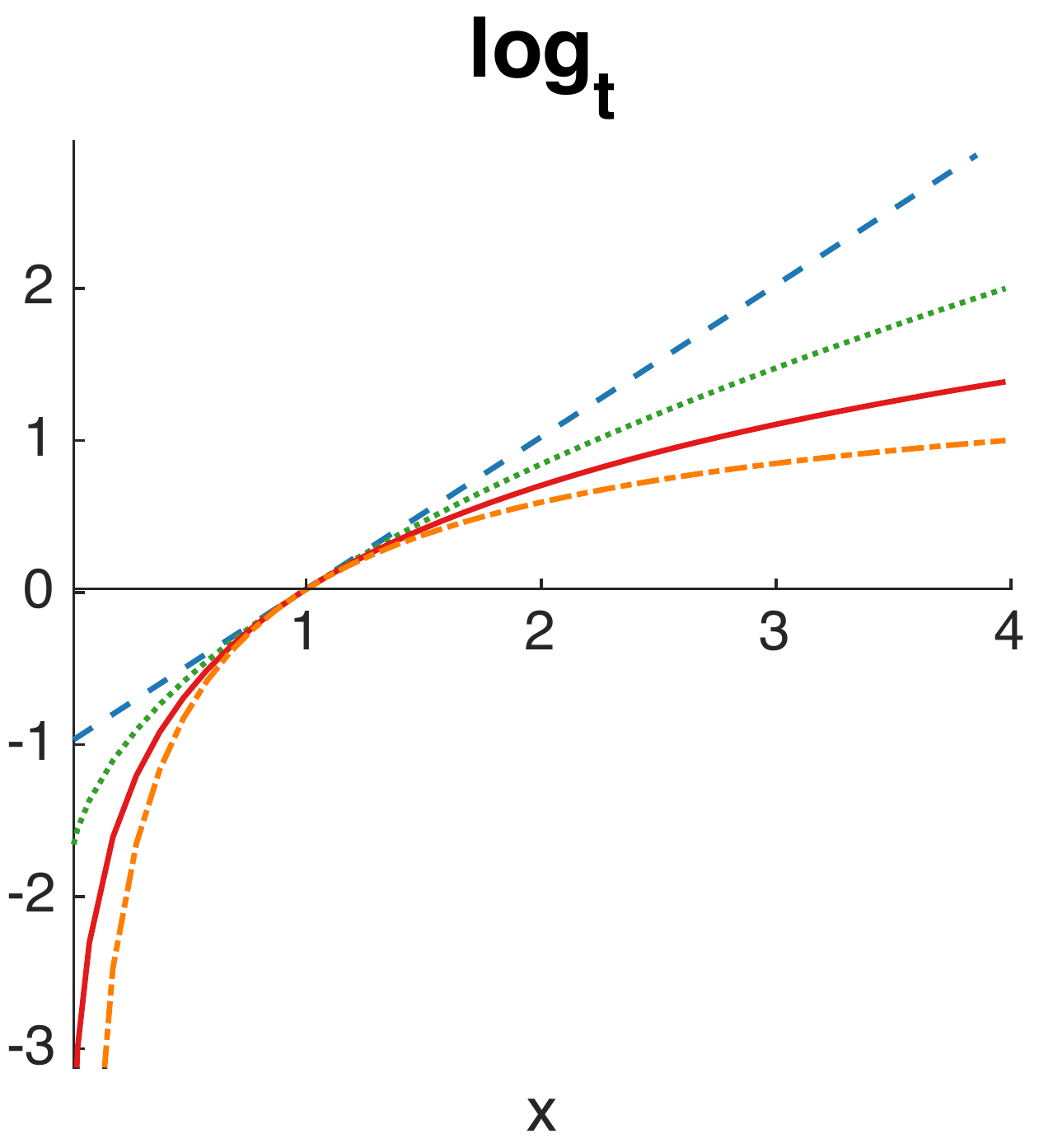}\label{fig:logt}}}\hfill
     \caption{Generalized $\exp$ and $\log$ functions: (a) $\exp_t$ function, and (b) $\log_t$ function for different values of $0 < t < 2$. Note that for $t =1$, the two functions reduce to standard $\exp$ and $\log$ functions, respectively.}\label{fig:texp}
\end{figure}

We first introduce the generalized $\log_t$ and $\exp_t$ functions as the generalization of the standard $\log$ and $\exp$ functions, respectively.
%
The generalized $\log_t$ function with \emph{temperature} parameter $0 < t < 2$ is defined as~\cite{texp1,texp3}
\begin{equation}
\label{eq:logt}
\log_t(x) = 
\begin{cases}
\log(x) & \text{if } t = 1\\
(x^{1-t}-1)/(1-t) & \text{otherwise}
\end{cases}\, .
\end{equation}
Note that $\log_t$ is concave and non-decreasing and generalizes the $\log$ function which is recovered in the limit $t \rightarrow 1$. The $\exp_t$ function is defined as the inverse of $\log_t$ function.
\begin{equation}
\label{eq:expt}
\exp_t(x) = 
\begin{cases}
\exp(x) & \text{if } t = 1\\
[1 + (1-t) x]_+^{1/(1-t)} & \text{otherwise}
\end{cases}\, ,
\end{equation}
where $[\,\cdot\,]_+ = \max(0,\cdot)$. Similarly, the standard $\exp$ is recovered in the limit $t \rightarrow 1$. Figure~\ref{fig:expt} and \ref{fig:logt} illustrate the $\exp_t$ and $\log_t$ functions for several values of $t$.

One major difference with the standard $\exp$ and $\log$ functions is that the familiar distributive properties do not hold in general: $\exp_t(a\,b) \neq \exp_t(a)\, \exp_t(b)$ and $\log_t(a\,b) \neq \log_t(a) + \log_t(b)$. An important property of $\exp_t$ is that it decays to zero slower than $\exp$ for values of $1 < t < 2$. This motivates defining heavy-tailed distributions using the $\exp_t$ function. More specifically, the $t$-exponential family of distributions is defined as a generalization of the exponential family by using the $\exp_t$ function in place of the standard $\exp$ function~\cite{texp4,sears}.

Our main focus is the \emph{capping} property of the $\log_t$ function: for values $x > 1$, the $\log_t$ function with $t > 1$ grows slower than the $\log$ function and reaches the constant value $1/(t-1)$ in the limit $x \rightarrow \infty$. This idea can be used to define the following robust loss transformation on the non-negative unbounded loss $\veps$:
\begin{equation}
\label{eq:type-II-ex}
\rho_t(\veps) = \log_t(1 + \veps),\quad 1 < t < 2\, .
\end{equation} 
Note that $\rho_t(0) = 0$, as desired. Moreover, the derivative of the transformed loss $\rho_t'(\veps) \rightarrow 0$ as $\veps \rightarrow \infty$ along with the additional property that the loss function converges to a constant as $\veps \rightarrow \infty$, i.e., $\rho_t(\veps) \rightarrow 1/(t-1) \geq 0$.
We will use this transformation to develop a robust ranking approach for the problem of the triplet embedding in presence of noise in the set of constraints.

Finally, note that setting $t = 1$ yields the transformation
\begin{equation}
\label{eq:type-I-ex}
\rho_1(\veps) = \log(1 + \veps)\, ,
\end{equation}
which has been used for robust binary ranking in~\cite{robirank}. Note that $\rho_1(\veps)$ grows slower than $\veps$, but still $\rho_1(\veps) \rightarrow \infty$ as $\veps \rightarrow \infty$. In other words, the transformed loss will not be capped from above. We will show that this transformation is not sufficient for robustness to noise.

\setcounter{subfigure}{-4}
\begin{figure*}[t]
 \vskip -0.1in
	\begin{center}
	\subfigure{\includegraphics[width=0.25\textwidth]{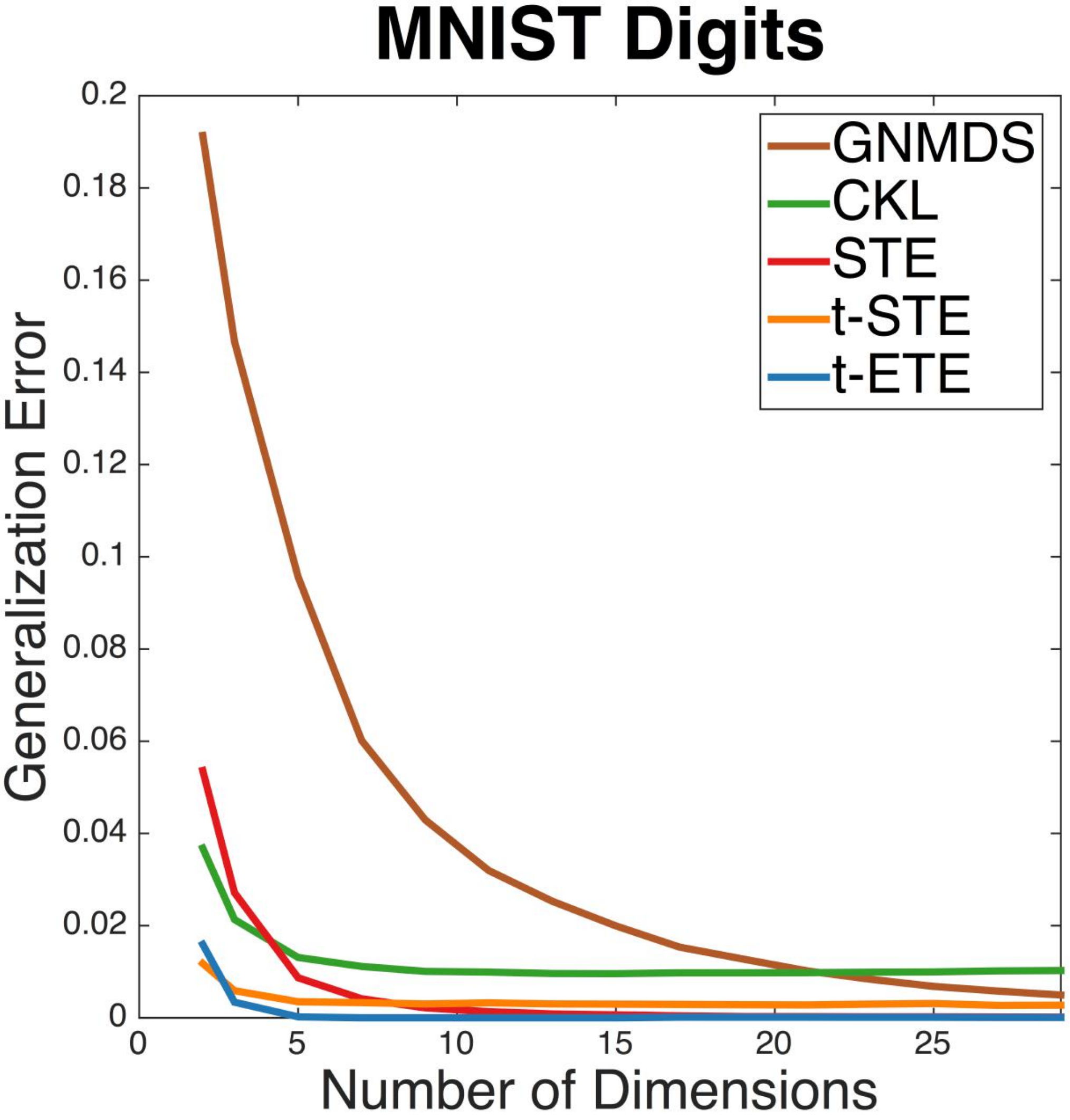}}\hfill
    \subfigure{\includegraphics[width=0.245\textwidth]{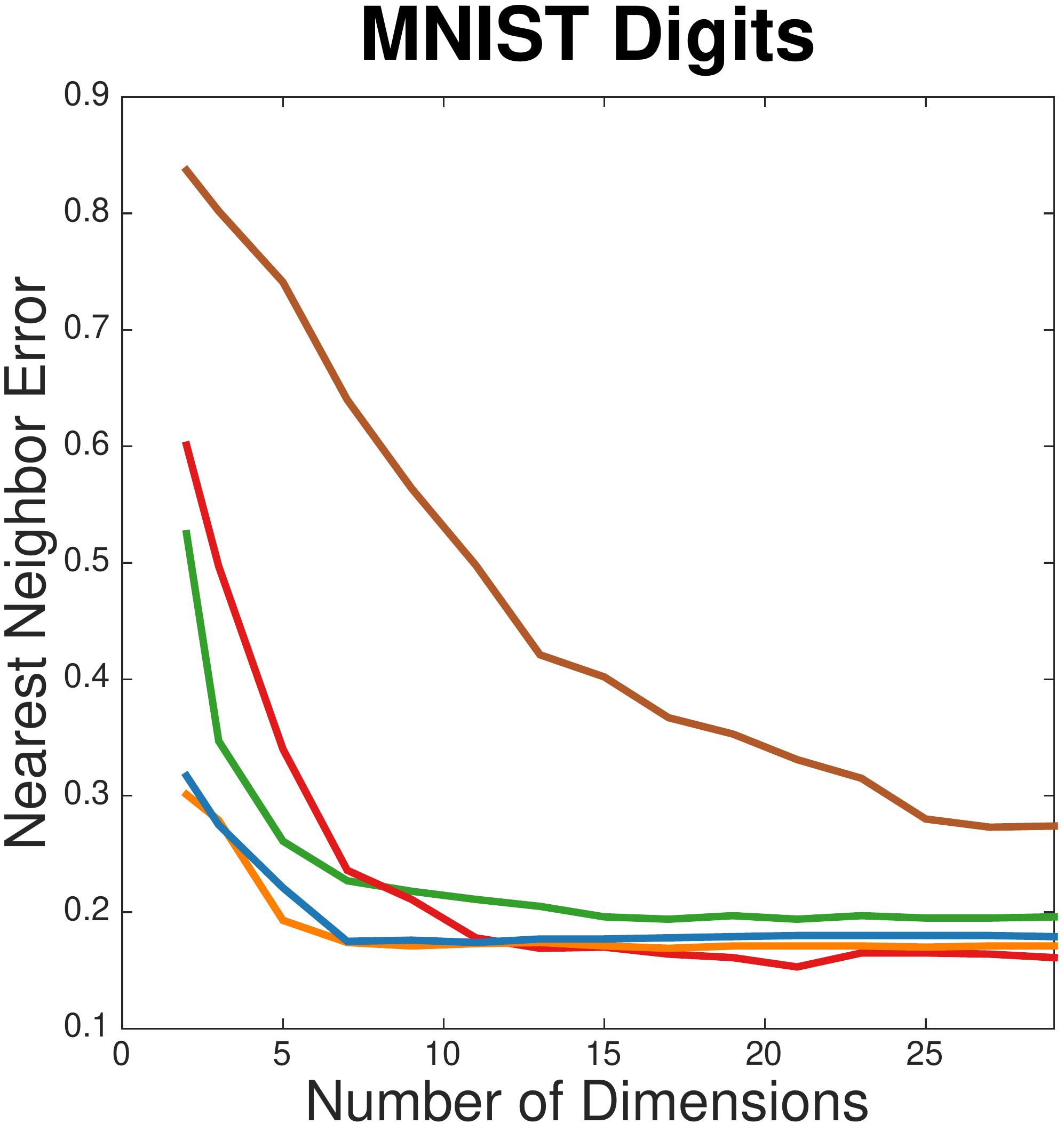}}\hfill
	\subfigure{\includegraphics[width=0.25\textwidth]{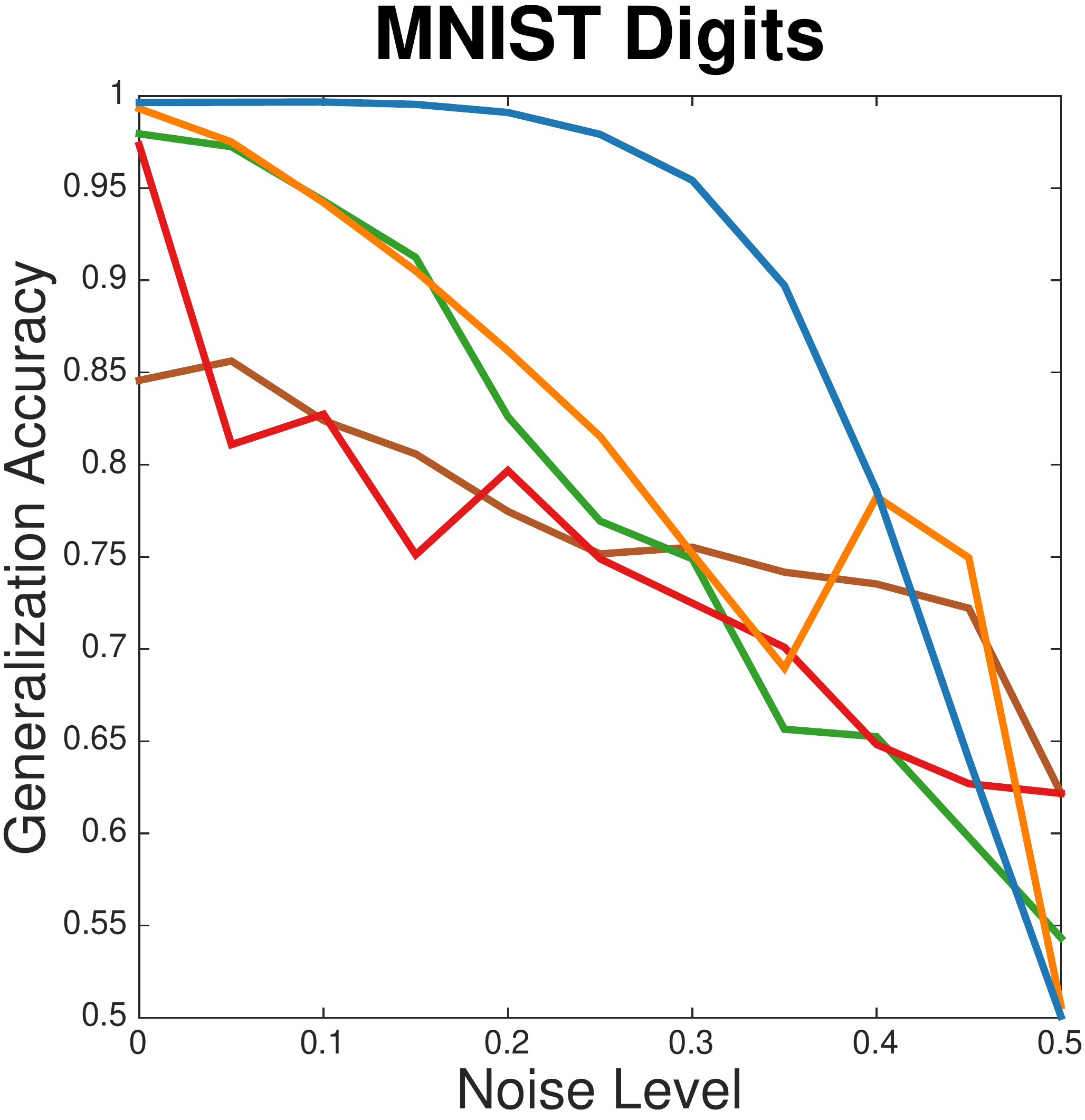}}\hfill
	\subfigure{\includegraphics[width=0.25\textwidth]{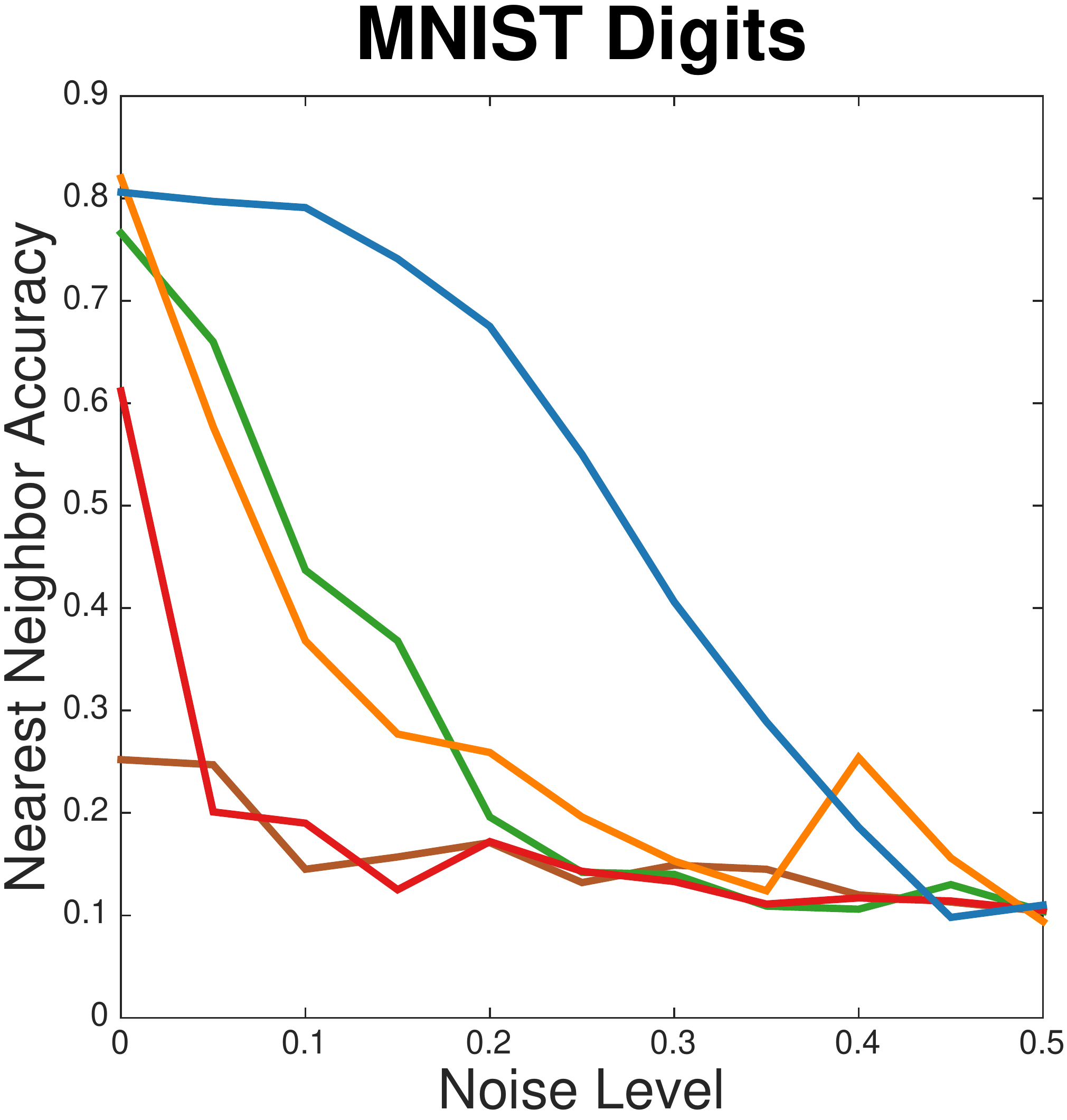}}\hfill\\\vskip -0.1in
	\subfigure[]{\includegraphics[width=0.25\textwidth]{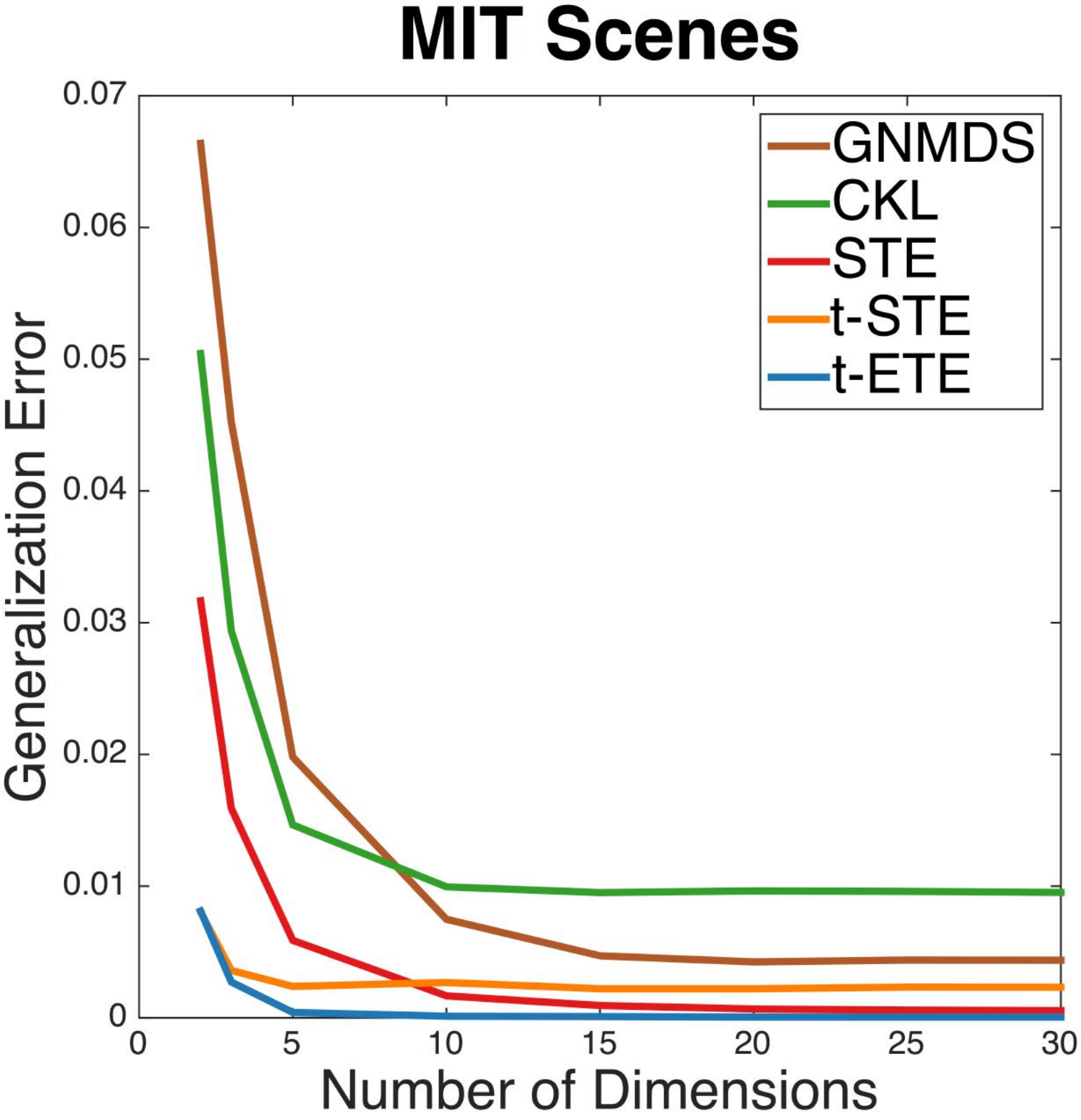}\label{fig:gen-err}}\hfill
    \subfigure[]{\includegraphics[width=0.25\textwidth]{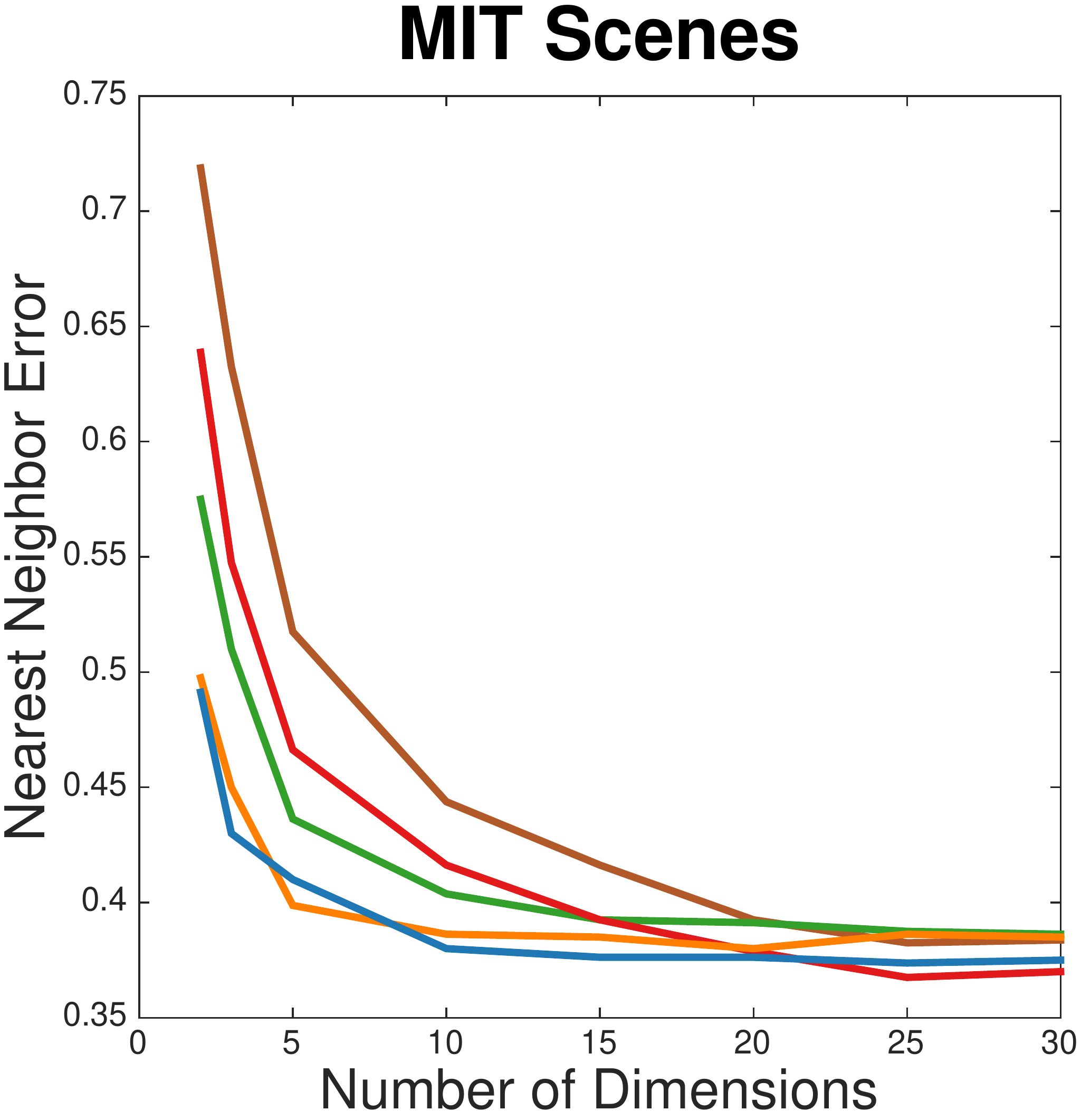}\label{fig:nn-err}}\hfill
	\subfigure[]{\includegraphics[width=0.245\textwidth]{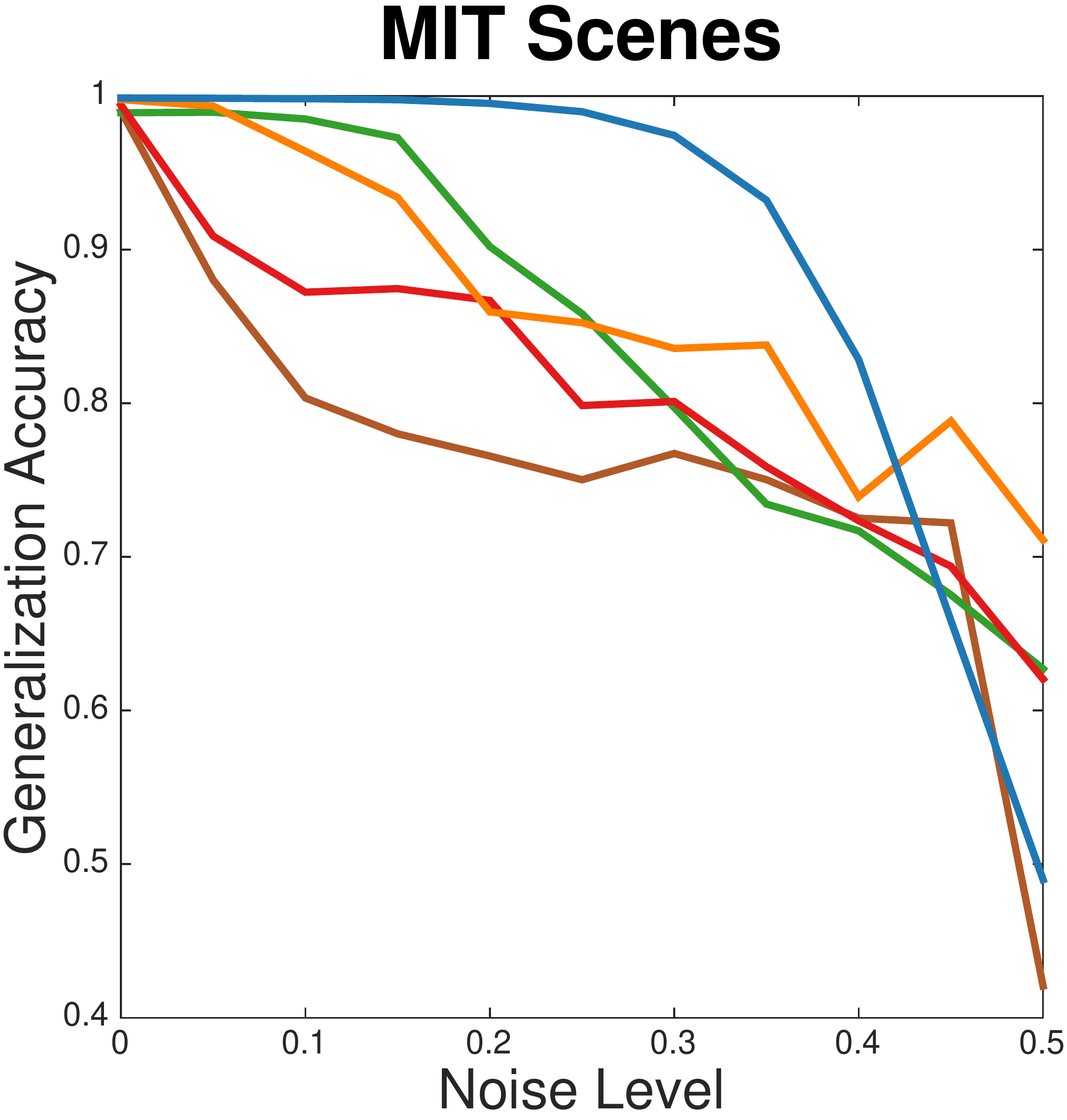}\label{fig:gen-err-noise}}\hfill
	\subfigure[]{\includegraphics[width=0.245\textwidth]{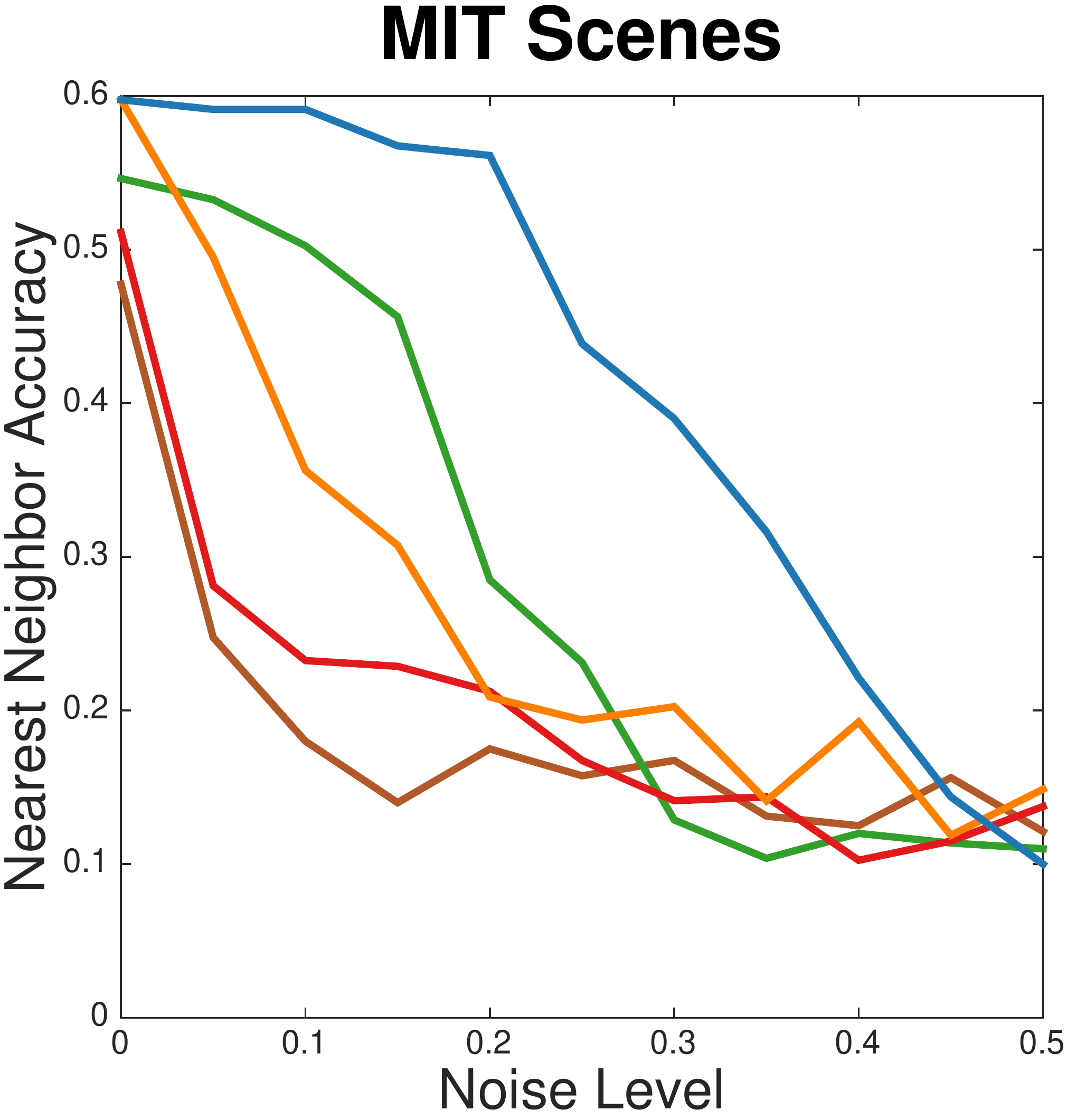}\label{fig:nn-err-noise}}\hfill
\caption{
Generalization and nearest-neighbor performance: \textbf{MNIST} (top row) and \textbf{MIT Scenes} (bottom row). (a) Generalization error, (b) nearest-neighbor error, (c) generalization accuracy in presence of noise, and (d) nearest-neighbor accuracy in presence of noise. For all the experiments, we use $t = t'$. For the generalization and nearest-neighbor error experiments, we start with $t = 2$ and use a smaller $t$ as the number of dimensions increases (more degree of freedom). For the noise experiments, we set $t = 1.7$. Figures best viewed in color.}\label{fig:results}
\end{center}
\end{figure*} 

\section{$t$-Exponential Triplet Embedding}
Building on our discussion on the heavy-tailed properties of generalized exp function~\eqref{eq:expt}, we can define the ratio
\begin{equation}
\label{eq:loss}
\tloss =  \frac{\exp_{t'}(-\Vert \y_i - \y_k\Vert^2)}{\exp_{t'}(-\Vert \y_i - \y_j\Vert^2)}
\end{equation}
with $1 < t' < 2$ as the loss of the ranking associated with the triplet $\triplet$. The loss is non-negative and satisfies the properties of a valid loss for ranking, as discussed earlier. 
Note that due to heavy-tail of $\exp_{t'}$ function with $1 < t' < 2$, the loss function~\eqref{eq:loss} encourages relatively higher-satisfaction of the ranking compared to, e.g., standard $\exp$ function.

Defining the loss of each triplet $\triplet \in \T$ as the ranking loss in~\eqref{eq:loss}, we formulate the objective of the triplet embedding problem as minimizing the sum of robust transformations of individual losses, that is,
\begin{align}
\label{eq:ete-obj}
\min_{\Y}\, \cLoss,\quad \cLoss  =\sum_{\triplet \in \T} \log_t\left(1 + \tloss\right)\, ,
\end{align}
in which, $1 < t < 2$. We call our method $t$-Exponential Triplet Embedding (\method, for short). Note that the loss of each triplet in the summation is now  capped from above by $1/(t-1)$. Additionally, the gradient of the objective function~\eqref{eq:ete-obj} with respect to the positions of the objects $\Y$
\begin{equation}
\label{eq:grad}
\nabla\cLoss = \sum_{\triplet \in \T} \frac{1}{(1 + \tloss)^t}\,\, \nabla\tloss
\end{equation}
includes additional \emph{forgetting factors} $1/(1 + \tloss)^t$ that damp the effect of those triplets that are highly-unsatisfied. 
\begin{algorithm}[tb]
   \caption{Weighted \method\ Dimensionality Reduction}
   \label{alg:wete}
\begin{algorithmic}
   \STATE {\bfseries Input:} high-dimensional data $\X = \{\x_1,\x_2,\ldots,\x_n\}$,\\ temperatures $t$ and $t^\prime$, embedding dimension $d$
   \STATE {\bfseries Output:}  $\Y = \{\y_1,\y_2,\ldots,\y_n\}$, where $\y_i \in \mathbb{R}^d$\\
   \medskip
   \STATE - $\T \leftarrow \{\}$, $\W \leftarrow \{\}$
   \FOR{$i=1$ {\bfseries to} $n$}
   		\FOR{$j$ {\bfseries} $\in \{m$-nearest neighbors of $i\}$}
 		\STATE - sample $k$ unif. from $\{k: \Vert \x_i - \x_k\Vert > \Vert \x_i - \x_j\Vert\}$
 		\STATE - compute weight $\omega_{ijk}$ using~\eqref{eq:weight}
 		\STATE - $\T \leftarrow \T \cup \triplet$
 		\STATE - $\W \leftarrow \W \cup \omega_{ijk}$
   		\ENDFOR
   \ENDFOR
   \STATE - for all $\omega \in \W$:  $\; \omega \leftarrow \frac{\omega}{\max_\omega \W} + \gamma$ 
   \STATE - initialize $\Y$ to $n$ points in $\mathbb{R}^d$\\ sampled from $\mathcal{N}(\mathbf{0},10^{-3}\mathbf{I}_{d\times d})$
   \FOR{$r = 1$ {\bfseries to} iter\#}
   \STATE - calculate the gradient $\nabla \cLossw$ of~\eqref{eq:wtete}
   \STATE - update $\Y \leftarrow \Y - \eta \nabla \cLossw$
   \ENDFOR
\end{algorithmic}
\end{algorithm}

%
%

\section{Connection to Previous Methods}
\label{sec:prev}

Note that by setting $t = 1$, we can use the property of the log function $\log (a) = -\log (1/a)$ to write the objective~\eqref{eq:ete-obj} as the following equivalent maximization problem\footnote{Note that $\log_t (a) \neq -\log_t (1/a)$ in general.}
\begin{equation}
\label{eq:obj-ste}
\max_{\Y} \sum_{\triplet \in \T} \log \pt\, ,
\end{equation}
where 
\begin{equation}
\label{eq:prob}
\pt = \frac{\exp_{t'}(-\Vert \y_i - \y_j\Vert^2)}{\exp_{t'}(-\Vert \y_i - \y_j\Vert^2) + \exp_{t'}(-\Vert \y_i - \y_k\Vert^2)}
\end{equation}
is defined as the \emph{probability} that the triplet $\triplet$ is satisfied. Setting $t' = 1$ and $t' = 2$ recovers the STE and t-STE (with $\alpha = 1$) formulations, respectively\footnote{The Student-t distribution with $\alpha$ degrees of freedom can be written in form of a $t$-exponential distribution with $-(\alpha+1)/2 = 1/(1-t)$ (see ~\cite{tlogreg}).}. STE (and t-STE) aim to maximize the joint probability that the triplets $\T$ are satisfied in the embedding $\Y$. The poor performance of STE and t-STE in presence of noise can be explained by the fact that there is no capping of the log-satisfaction probabilities\footnote{Note that in this case, the probabilities should be capped from below.} of each triplet (see~\eqref{eq:type-I-ex}). Therefore, the low satisfaction probabilities of a the noisy triplet dominates the objective function~\eqref{eq:obj-ste} and thus, results in poor performance.

\begin{figure*}[t!]
	\begin{center}
    \subfigure[]{\raisebox{1cm}[0pt][0pt]{\includegraphics[width=0.33\textwidth]{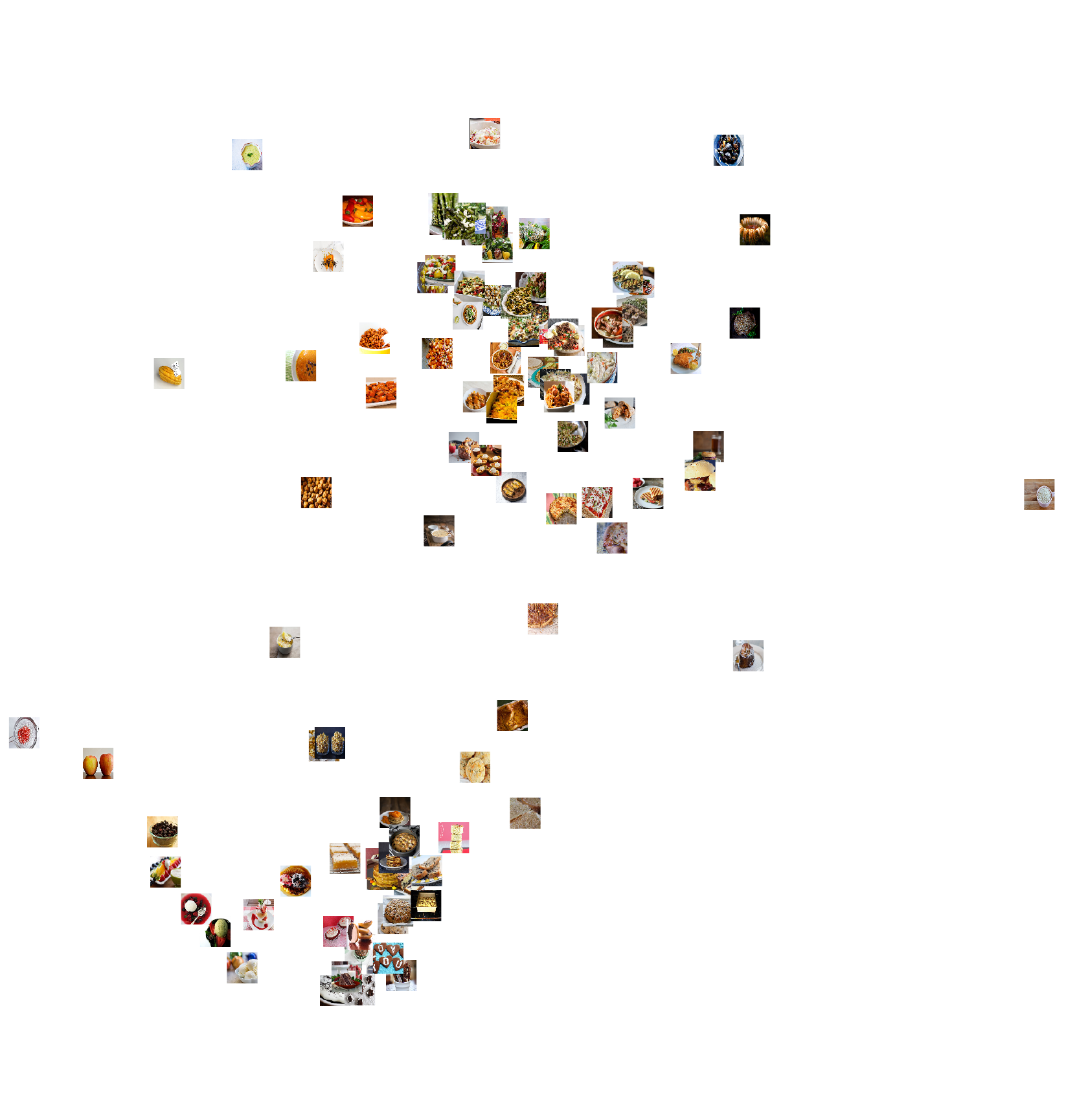}}\label{fig:food-tste}} \hspace{1.5cm}
	\subfigure[]{\includegraphics[width=0.33\textwidth]{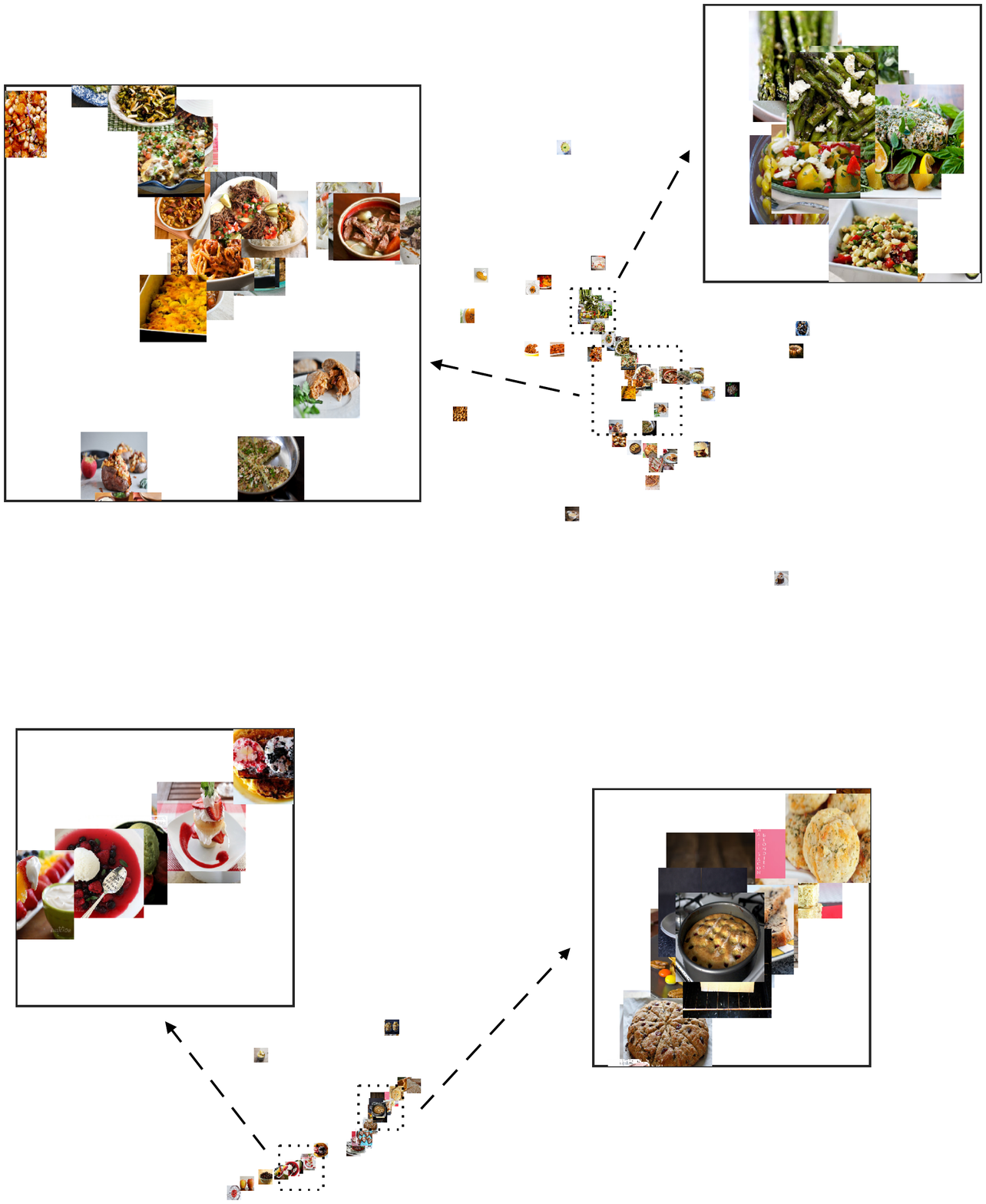}\label{fig:food-tete}}
\caption{\label{fig:food}
Embedding of the \textbf{Food} dataset using (a) t-STE, and $t$-ETE ($t = 2$) methods. There appear no clear separation between the clusters in (a) while in (b), three different clusters of food are evident: ``Vegetables and Meals'' (top), ``Ice creams and Deserts'' (bottom left), and ``Breads and Cookies'' (bottom right).}
\end{center}
\end{figure*} 

\section{Applications to Dimensionality Reduction}
\label{sec:dr}

Now, consider the case where a high-dimensional representation $\X = \{\x_i\}_{i=1}^n$ is provided for a set of $n$ objects. Having the  $t$-ETE method in hand, one may ask the following question: ``given the high-dimensional representation $\X$ for the objects, is it possible to find a lower-dimensional representation $\Y$ for these objects by satisfying a set of ranking constraints (i.e., triplets), formed based on their relative similarities in the representation $\X$?''. Note that the total number of triplets that can be formed on a set of $n$ objects is $\mathcal{O}(n^3)$ and trying to satisfy all the possible triplets is computationally expensive. 
However, we argue that most of the these triplets are redundant and contain the same amount of information about the relative similarity of the objects. For instance, consider two triplets $\triplet$ and $(i,j, k')$ in which $i$ and $k$ are located far away and $k$ and $k'$ are neighbors of each other. Given $\triplet$, having $(i,j, k')$ provides no extra information on the placements of $i$ and $j$, as long as $k$ and $k'$ are located close together in the embedding. In other words, $k$ and $k'$ are viewed by $i$ as almost being the same object.

Note that for each object $i$, the nearby objects having relatively short distance to $i$ specify the local structure of the object, whereas those that are located far away determine the global placement of $i$ in the space. For that matter, for each query object $i$, we would like to consider those triplets (with high probability) that preserve both local and global structure of the data. Following the discussion above, we emphasize on preserving the local information by explicitly choosing the first test object among the nearest-neighbors of the query object $i$. The global information of the object $i$ is then preserved by considering a small number of objects, uniformly sampled from those that are located farther away. This leads to the following procedure for sampling a set of informative triplets. For each object $i$, we choose the first object from the set of $m$-nearest neighbors of $i$ and then, sample the outlier object uniformly from those that are located farther away from $i$ than the first object. This is equivalent to sampling a triplet uniformly at random conditioned on that the first test object is chosen among the $m$-nearest neighbors of $i$. We use equal number of nearest-neighbors and outliers for each point, which results in $nm^2$ triplets in total.

The original $t$-ETE formulation aims to satisfy each triplet equally likely. This would be reasonable in cases where no side information about the extent of each constraint is provided. However, given the high-dimensional representation of the objects $\X$, this assumption may not be accurate. In other words, the ratio of the pairwise similarities of the objects specified in each triplet may vary significantly among the triplets. To account for this variation, we can introduce a notion of weight for each triplet to reflect the extent that the triplet needs to be satisfied. More formally, let $\omega_{ijk} \geq 0$ denote the weight associated with the triplet $\triplet$ and let $\W = \{\omega_{ijk}\}$ denote the set of all triplet weights. The Weighted $t$-ETE can be formulated as minimizing the sum of weighted capped losses of triplets, that is,
\begin{equation}
\label{eq:wtete}
\min_{\Y}\,\cLossw, \quad \cLossw = \sum_{\triplet \in \T} \omega_{ijk} \log_t\left(1 + \tloss\right).
\end{equation}
The $t$-ETE method can be seen as a special case of the weighted triplet embedding formulation where all the triplets have unit weights.

\begin{figure*}[th]
	\begin{center}
	\includegraphics[width=0.9\textwidth]{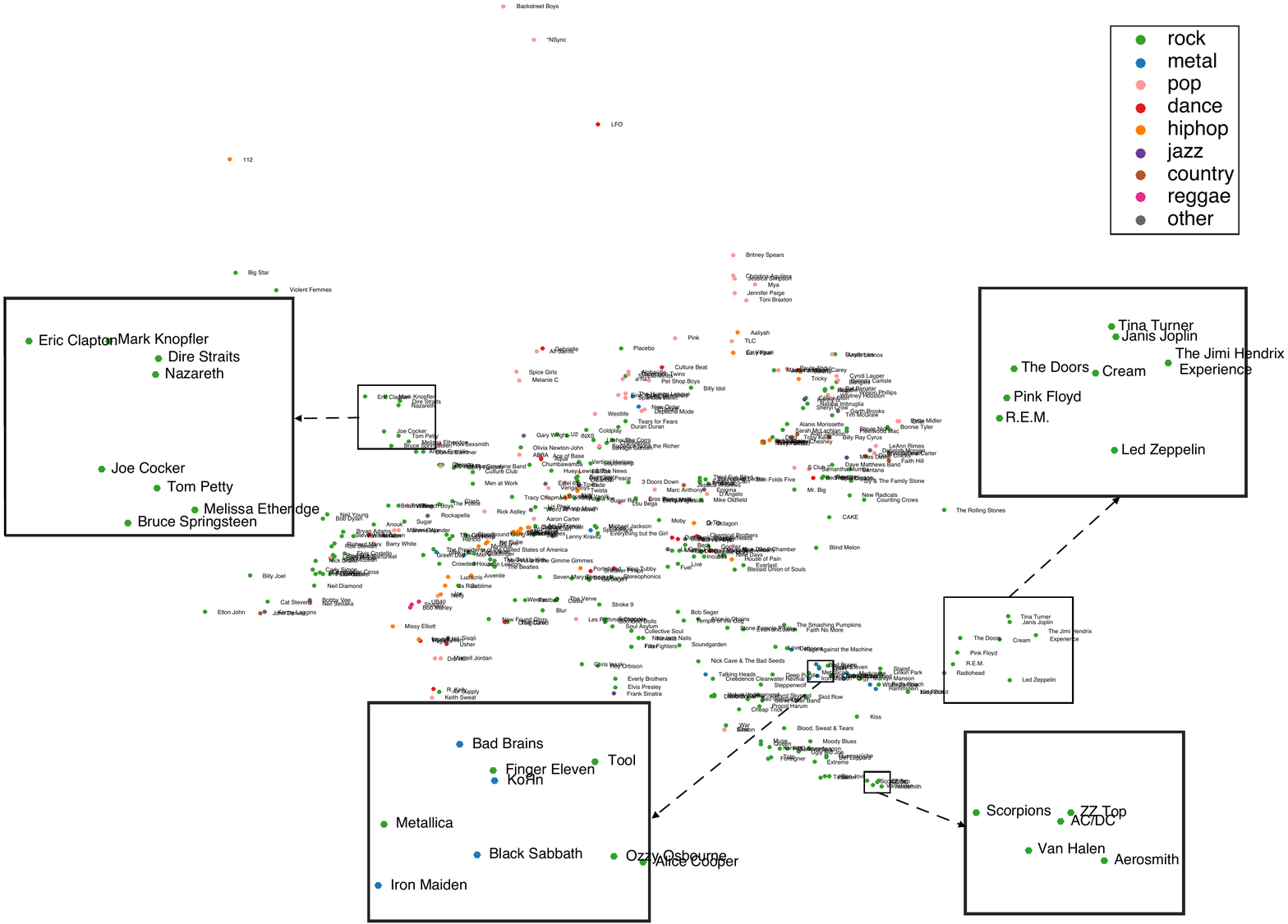}
\caption{Results of the $t$-ETE algorithm ($t = 2$) on the \textbf{Music} dataset: compare the result with the one in~\cite{ste}. The neighborhood structure is more meaningful in some regions than the one with t-STE.}\label{fig:music}
\end{center}
\end{figure*} 

Finally, to assign weights to the sampled triplets, we note that the loss ratio in~\eqref{eq:loss} is inversely proportional to how well the triplet is \emph{satisfied} in the embedding. This suggests using the inverse loss ratios of the triplets in the high-dimensional space as the weights associated with the triplets. More formally, we set
\begin{equation}
\label{eq:weight}
\omega_{ijk} = \frac{\exp(-\Vert \x_i - \x_j\Vert^2/\sigma^2_{ij})}{\exp(-\Vert \x_i - \x_k\Vert^2/\sigma^2_{ik})}\, ,
\end{equation}
 where $\sigma^2_{ij} = \sigma_i\, \sigma_j$ is a constant scaling factor for the pair $(i,j)$. We set $\sigma_i$ to the distance of $i$ to its $10$-th nearest neighbor. This choice of scaling adaptively handles the dense as well as the sparse regions of data distribution. Finally, the choice of $\exp$ function rather than using $\exp_{t'}$ with $1< t' < 2$ is to have more emphasis on the distances of the objects in the high-dimensional space. The pseudo-code for the algorithm is shown in Algorithm~\ref{alg:wete}. In practice, dividing each weight at the end by the maximum weight in $\W$ and adding a constant positive bias $\gamma > 0$ to all weight improves the results.

Note that both sampling and weighting the triplets using~\eqref{eq:weight} and calculating the gradient of loss requires calculating the pairwise distances only between $\mathcal{O}(nm^2)$ objects in the high-dimensional space (for instance, by using efficient methods to calculate $m$-nearest neighbors such as~\cite{knn}) or the low-dimensional embedding. In many cases, $m^2 \ll n$, which results in a huge computational advantage over $\mathcal{O}(n^2)$ complexity of t-SNE.

\section{Experiments}
\label{sec:exp}

In this section, we conduct experiments to evaluate the performance of  $t$-ETE for triplet embedding as well the application of Weighted $t$-ETE for non-linear dimensionality reduction. In the first set of experiments, we compare $t$-ETE to the following triplet embedding methods: $1$) GNMDS, $2$) CKL, $3$) STE, and $4$) t-STE. We evaluate the generalization performance of the different methods by means of satisfying unseen triplets and the nearest-neighbor error, as well as their robustness to constraint noise. We also provide visualization results on two real-world datasets. Next, we apply the Weighted $t$-ETE method for non-linear dimensionality reduction and compare the result to the t-SNE method. The code for the (Weighted) $t$-ETE method as well as all the experiments will be publicly available upon acceptance. 

%

\begin{figure*}[th!]
\vspace{0cm}
\begin{center}
\centering\sf\setlength{\tabcolsep}{57pt}
\begin{tabular}{|*{4}{C|}}
\hline
   \includegraphics[width=0.20\textwidth]{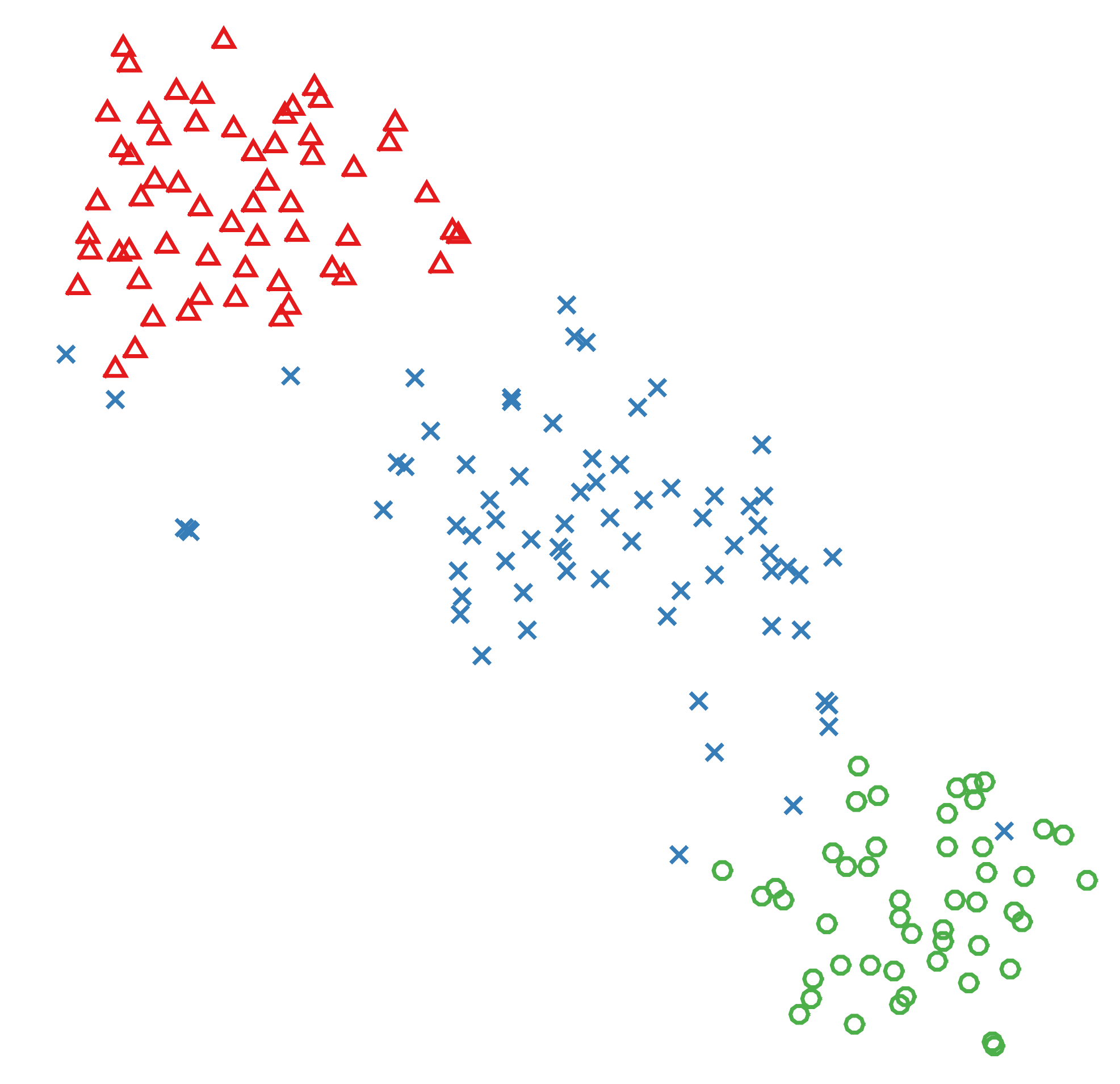}
& \raisebox{0.3cm}[0pt][0pt]{\includegraphics[width=0.20\textwidth]{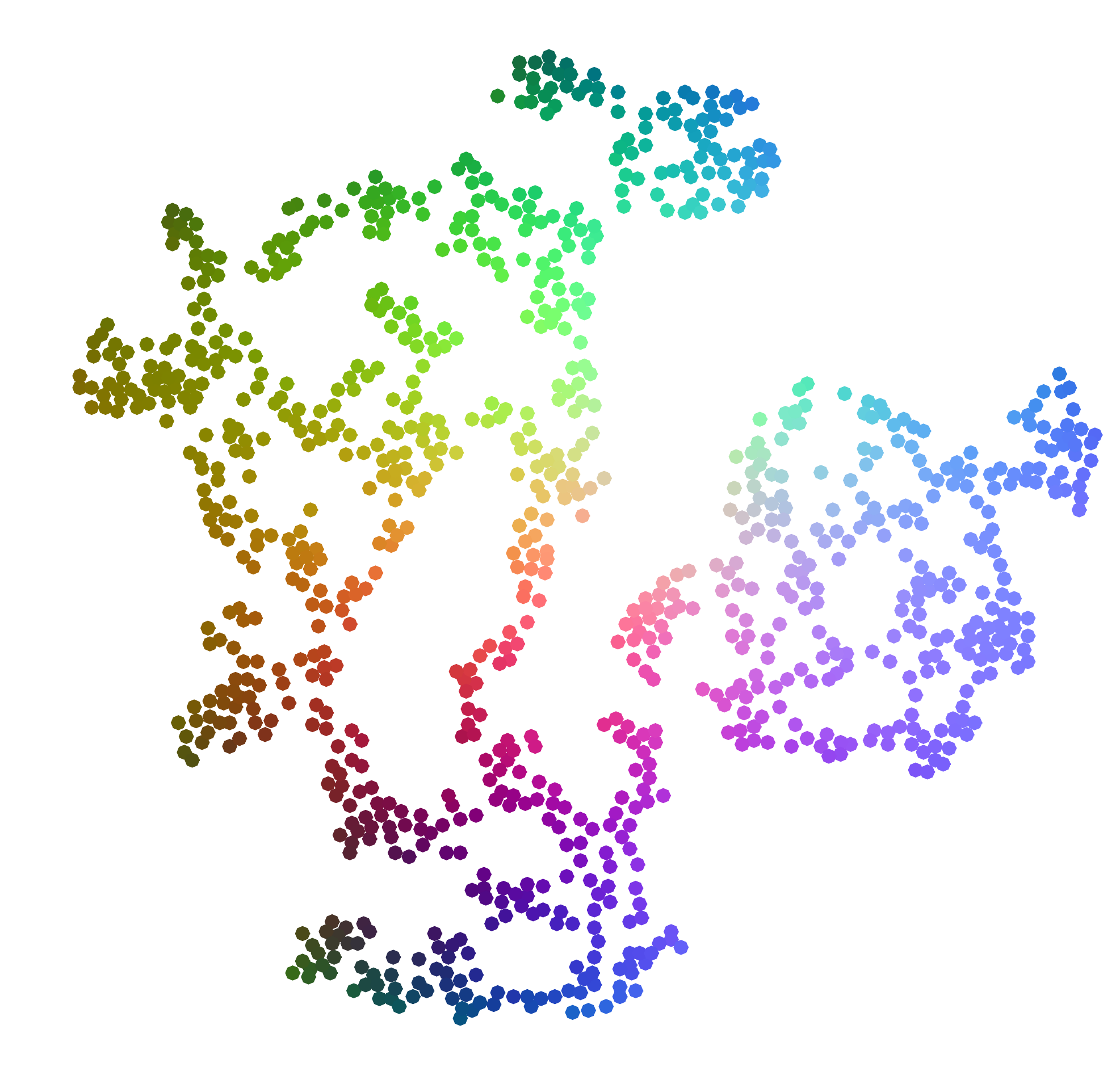}}
& \raisebox{0.3cm}[0pt][0pt]{\includegraphics[width=0.20\textwidth]{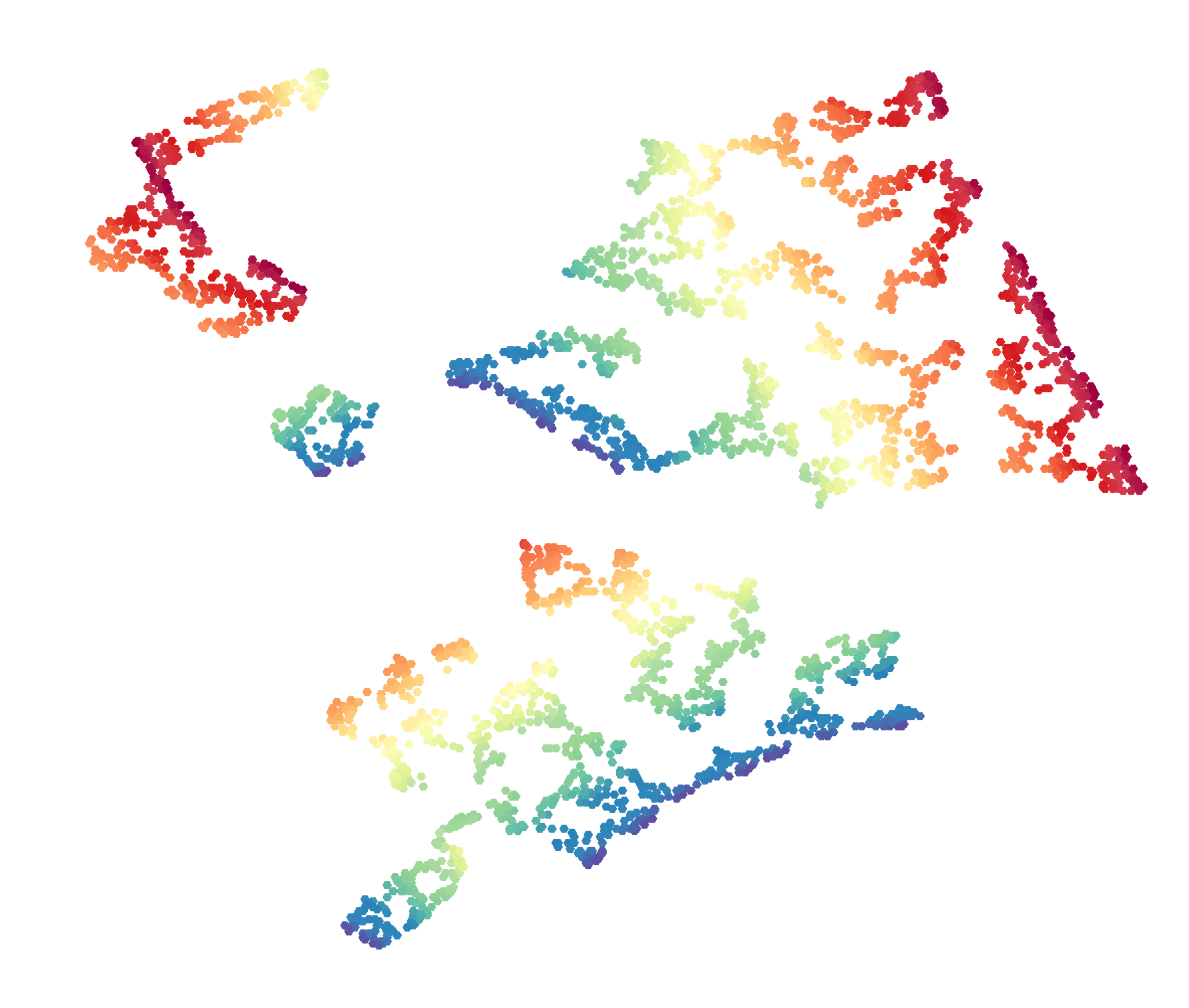}}
& \includegraphics[width=0.20\textwidth]{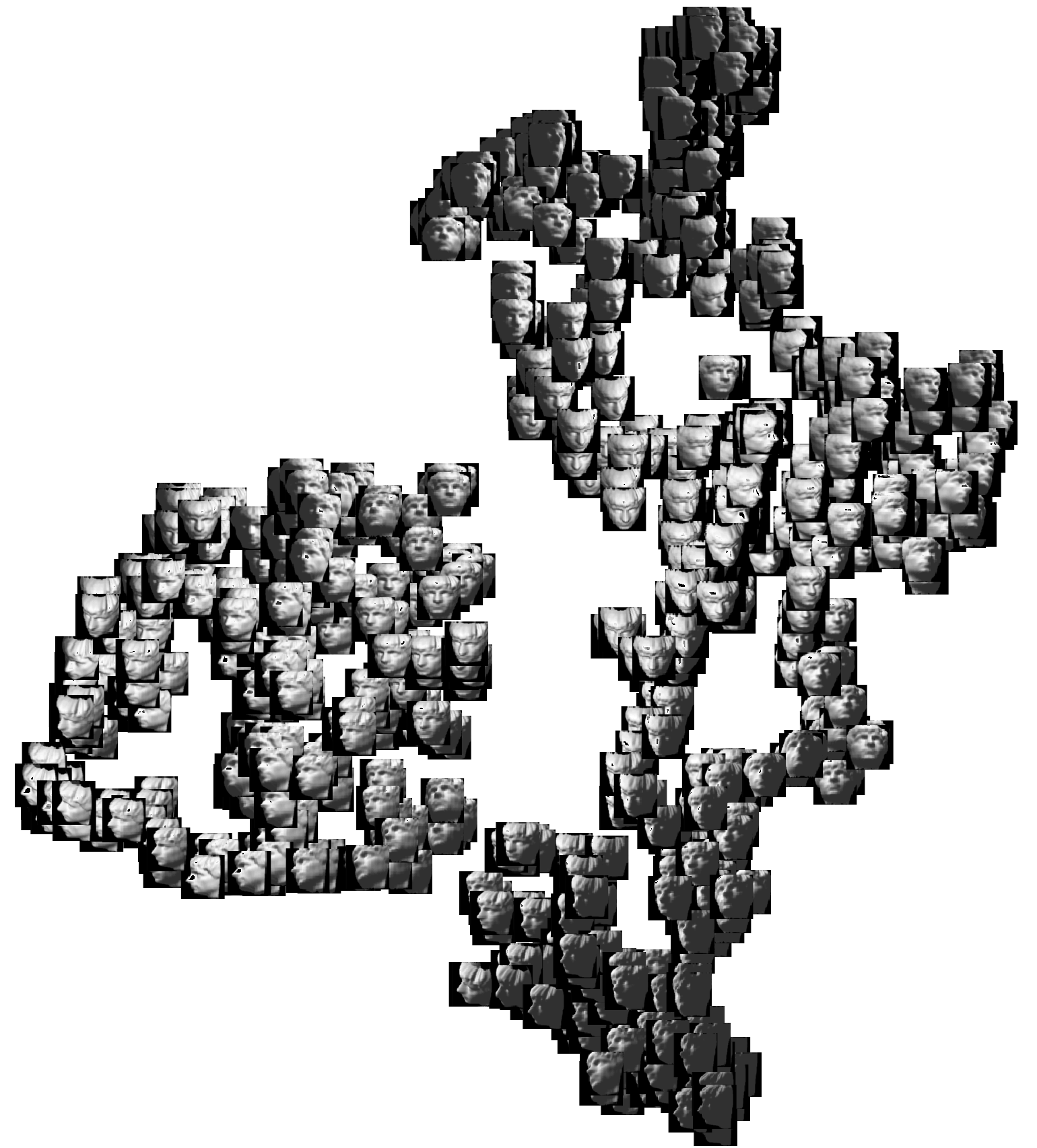}\\\hline
   \includegraphics[width=0.20\textwidth]{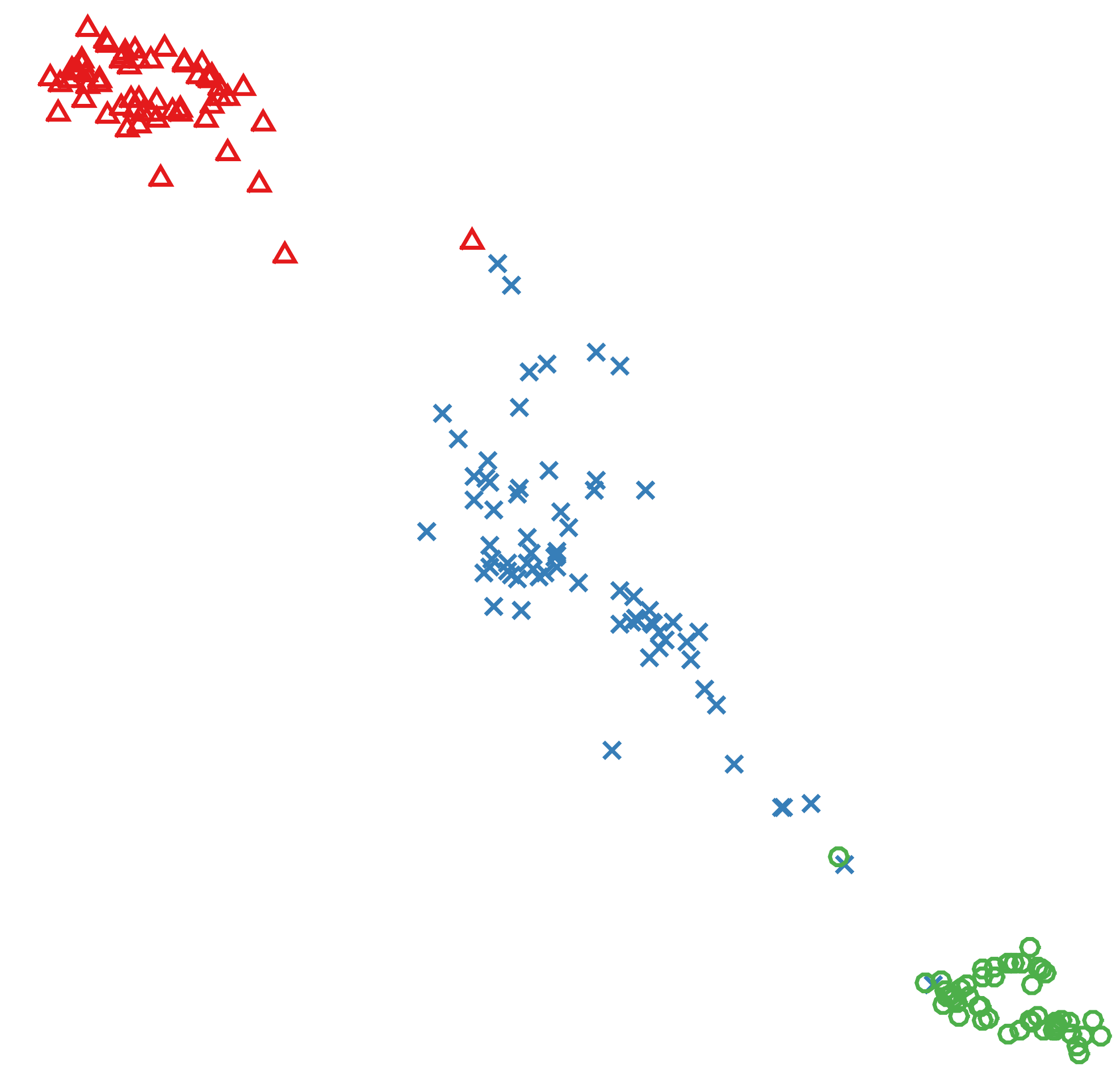}
& \raisebox{0.3cm}[0pt][0pt]{\includegraphics[width=0.20\textwidth]{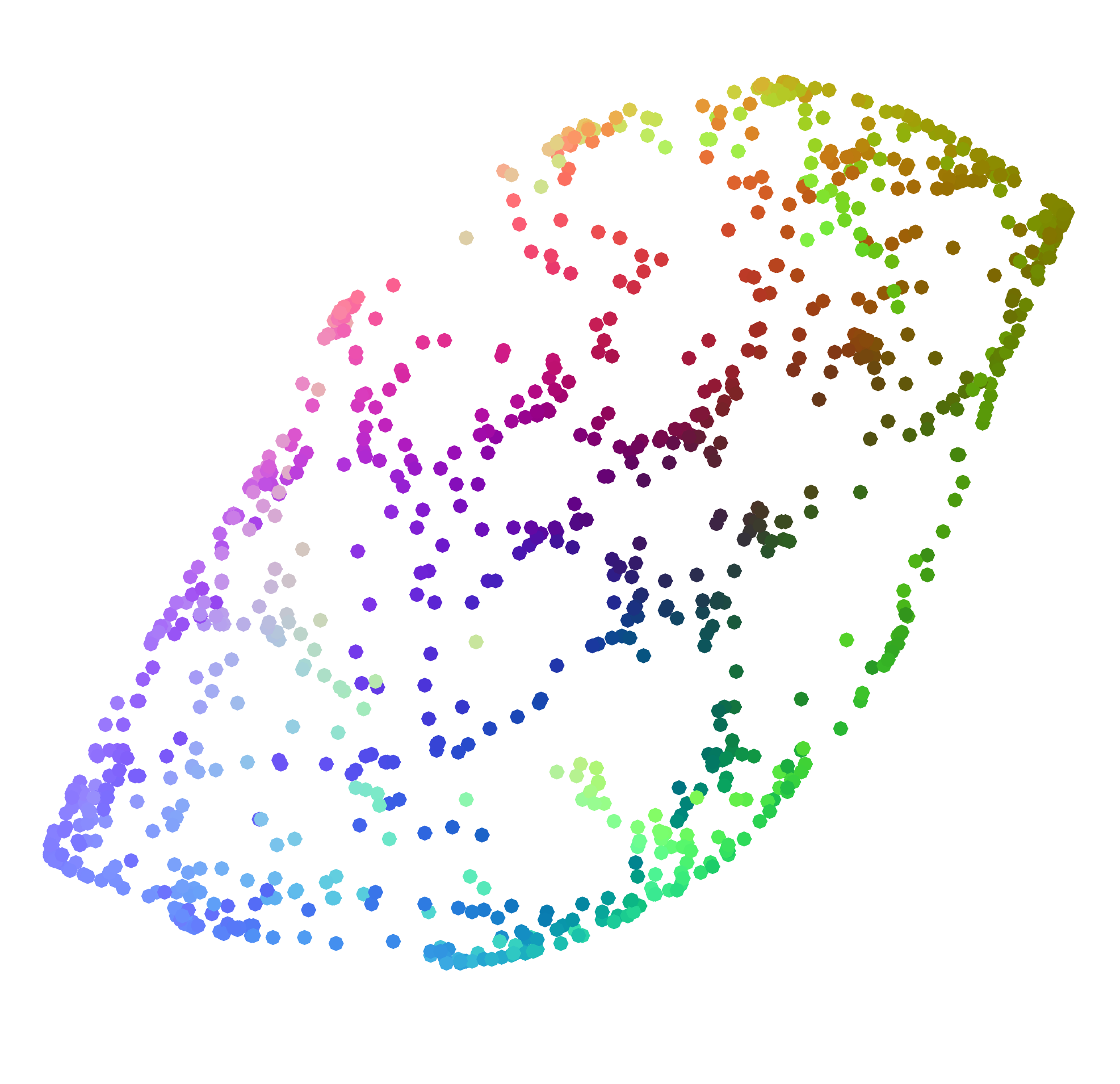}}
& \raisebox{0.3cm}[0pt][0pt]{\includegraphics[width=0.20\textwidth]{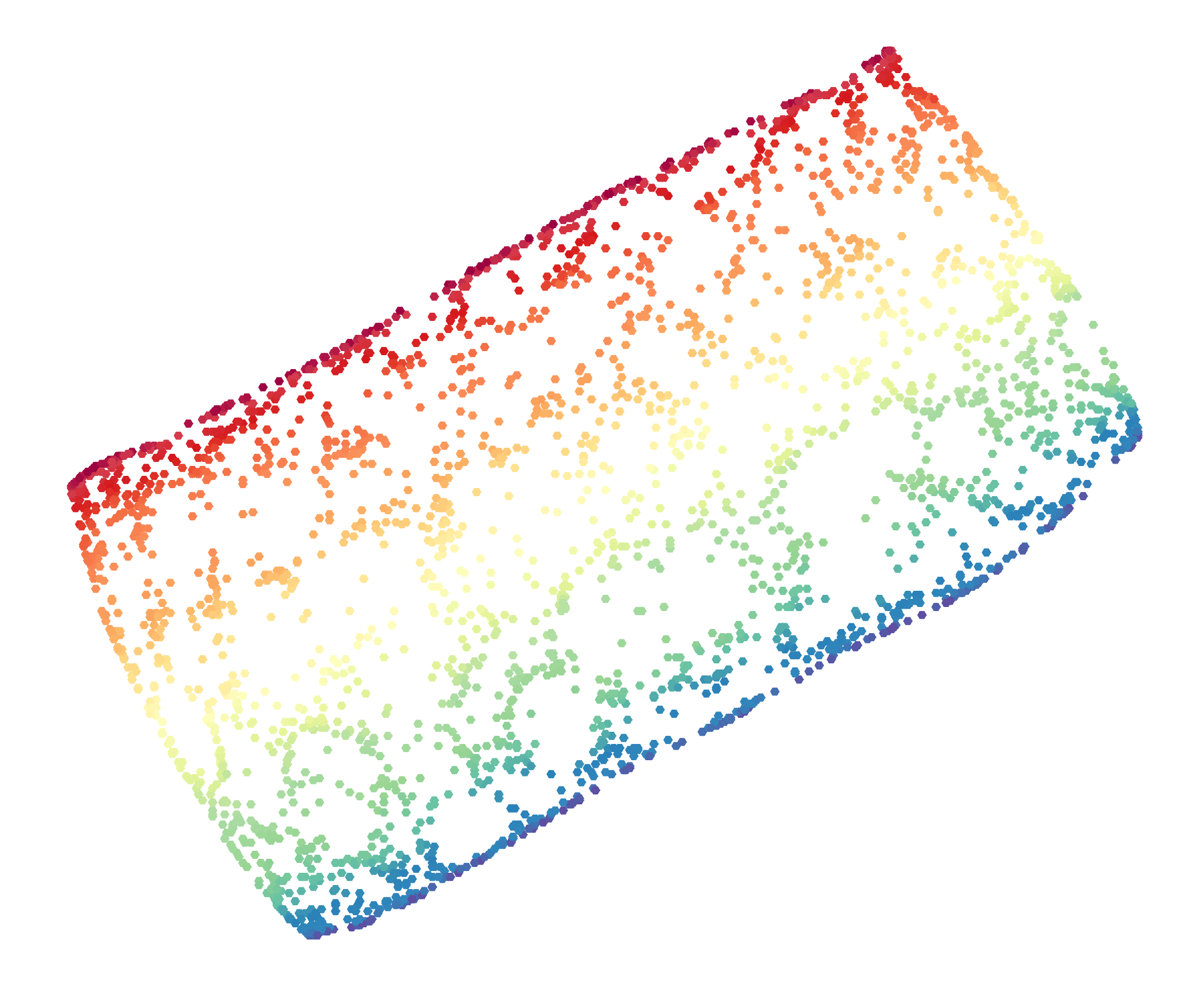}}
& \includegraphics[width=0.20\textwidth]{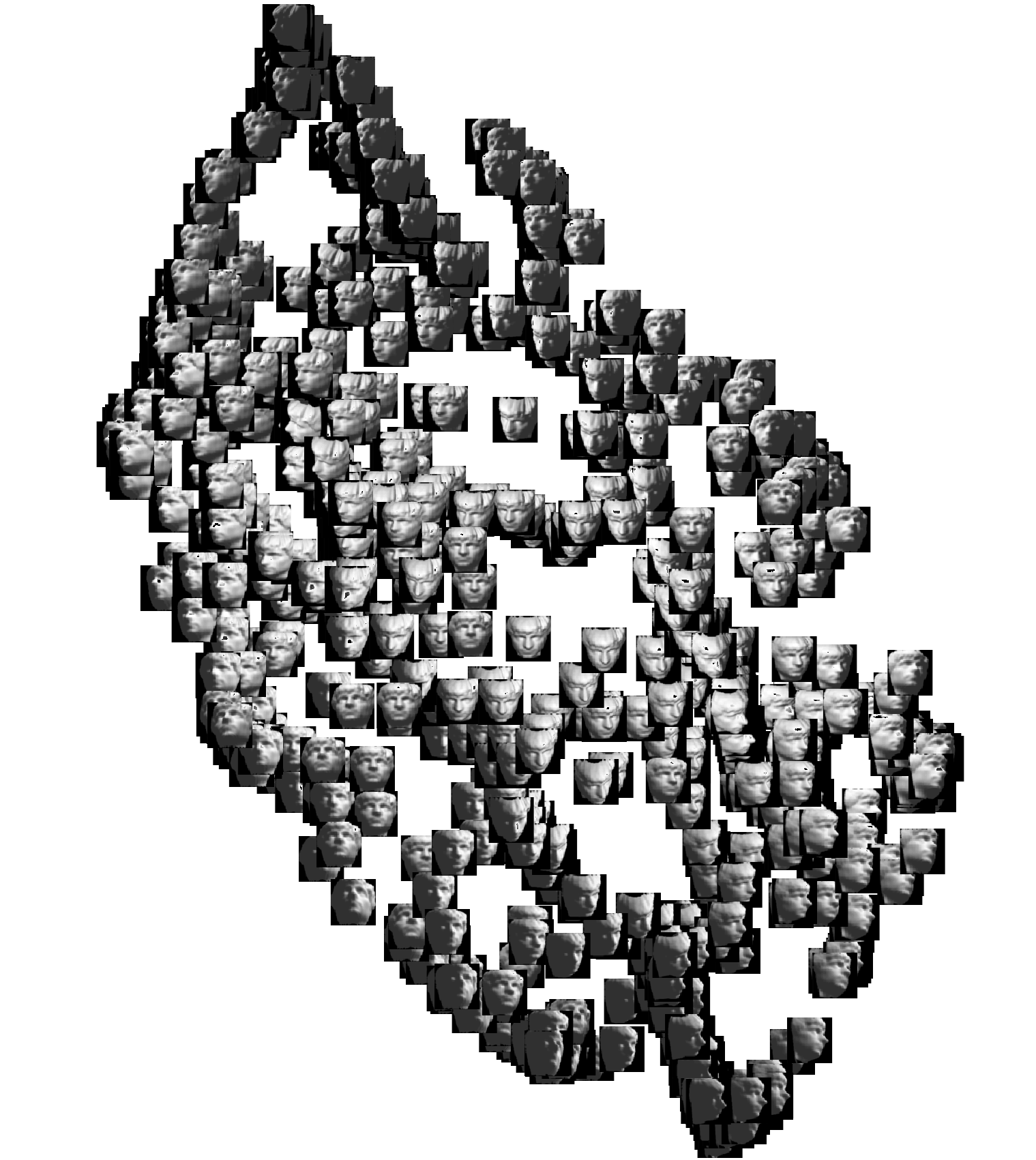}\\\hline
    \multicolumn{1}{c}{(a)} & \multicolumn{1}{c}{(b)} & \multicolumn{1}{c}{(c)} & \multicolumn{1}{c}{(d)}\\\hline
   \includegraphics[width=0.22\textwidth]{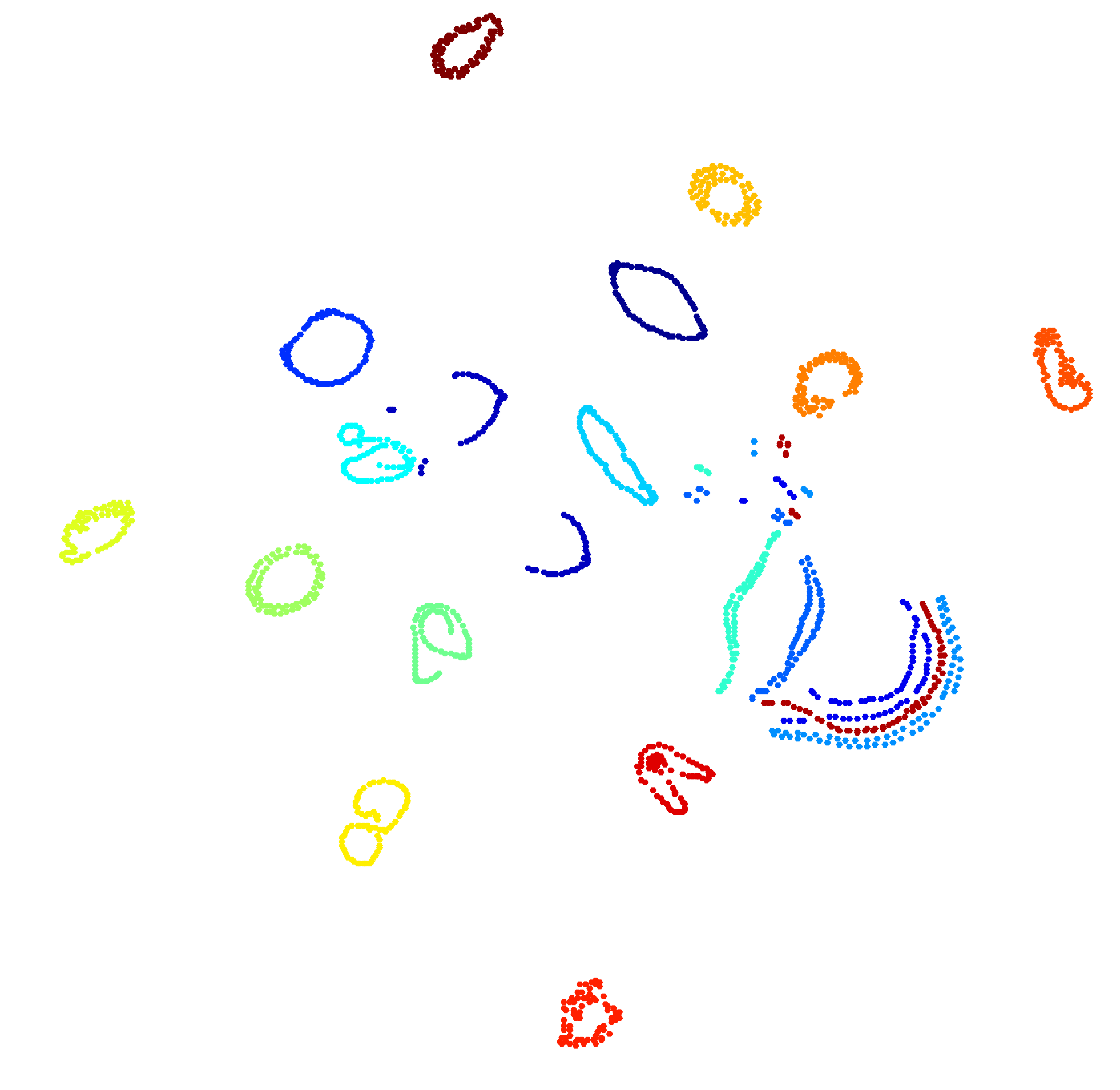}
& \includegraphics[width=0.24\textwidth]{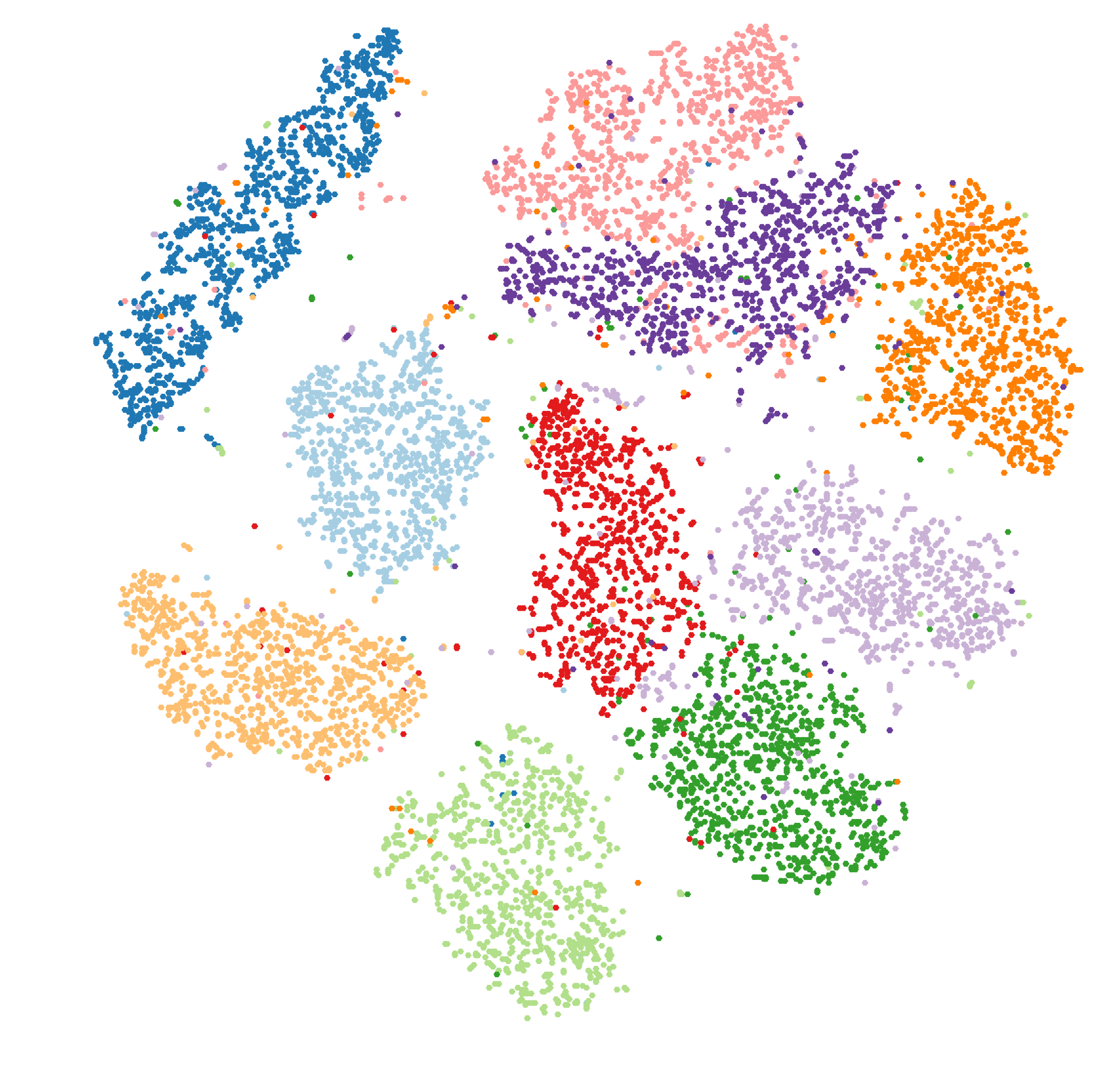}
& \includegraphics[width=0.22\textwidth]{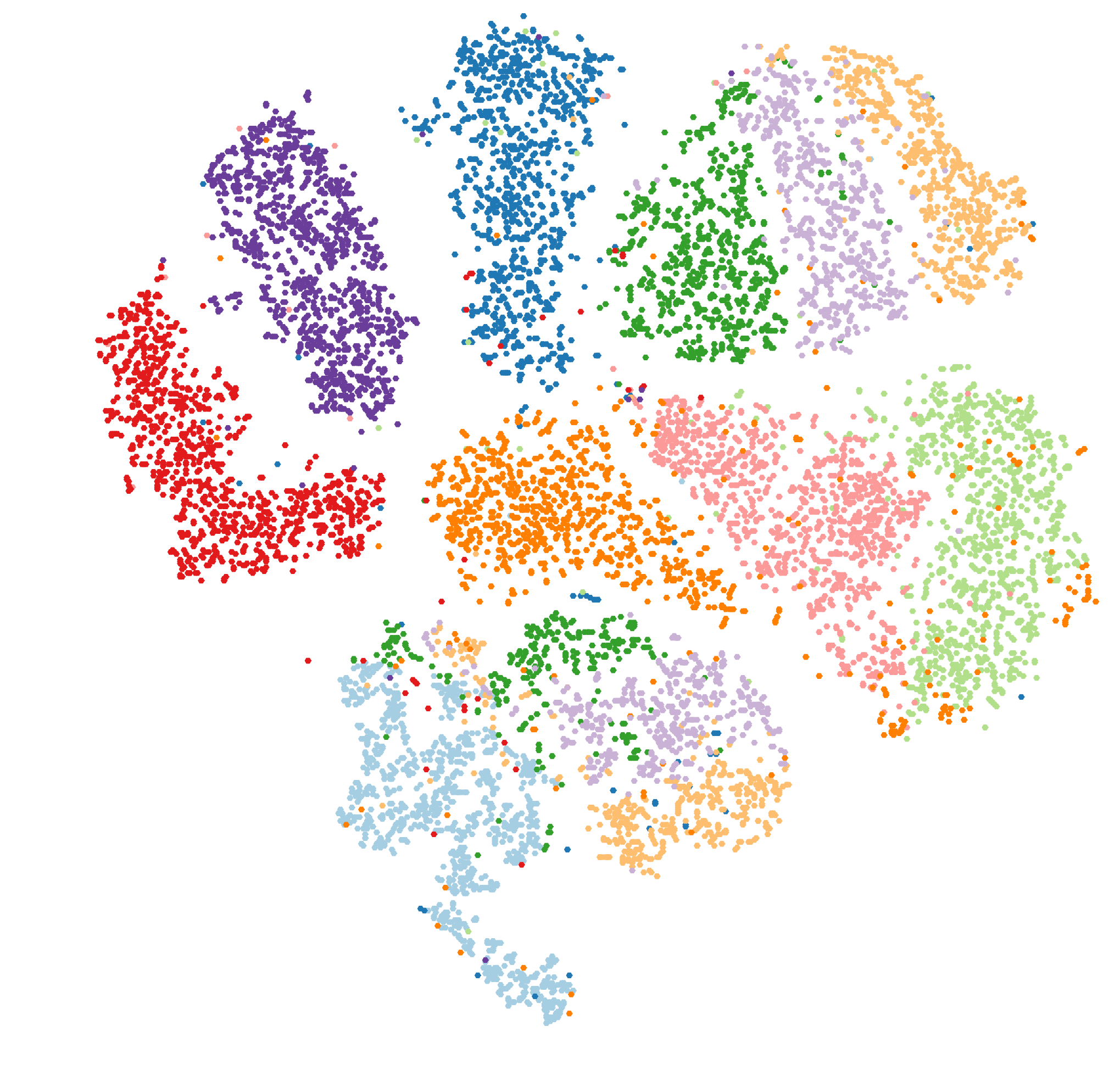}
& \includegraphics[width=0.22\textwidth]{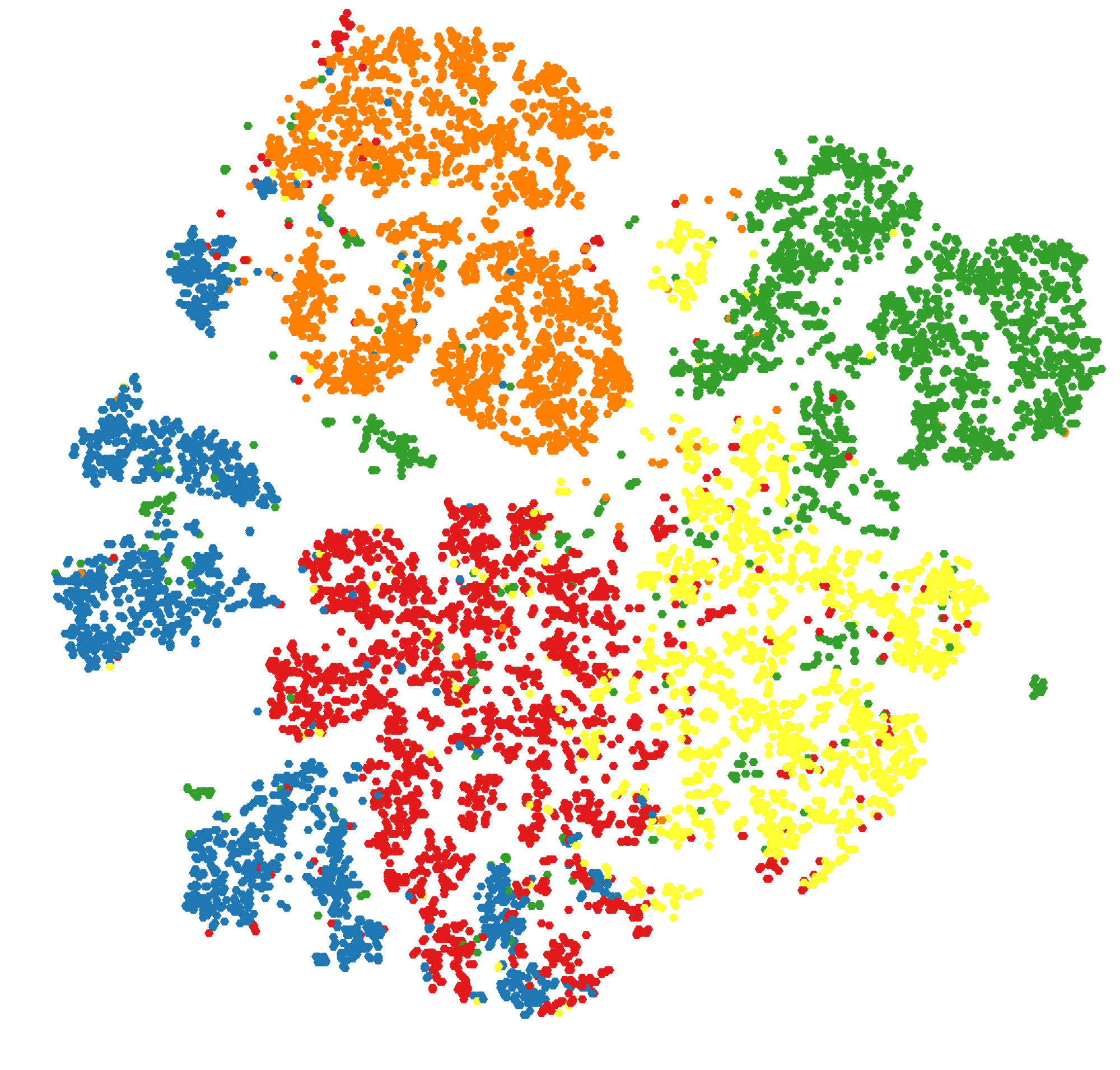}\\\hline
   \includegraphics[width=0.22\textwidth]{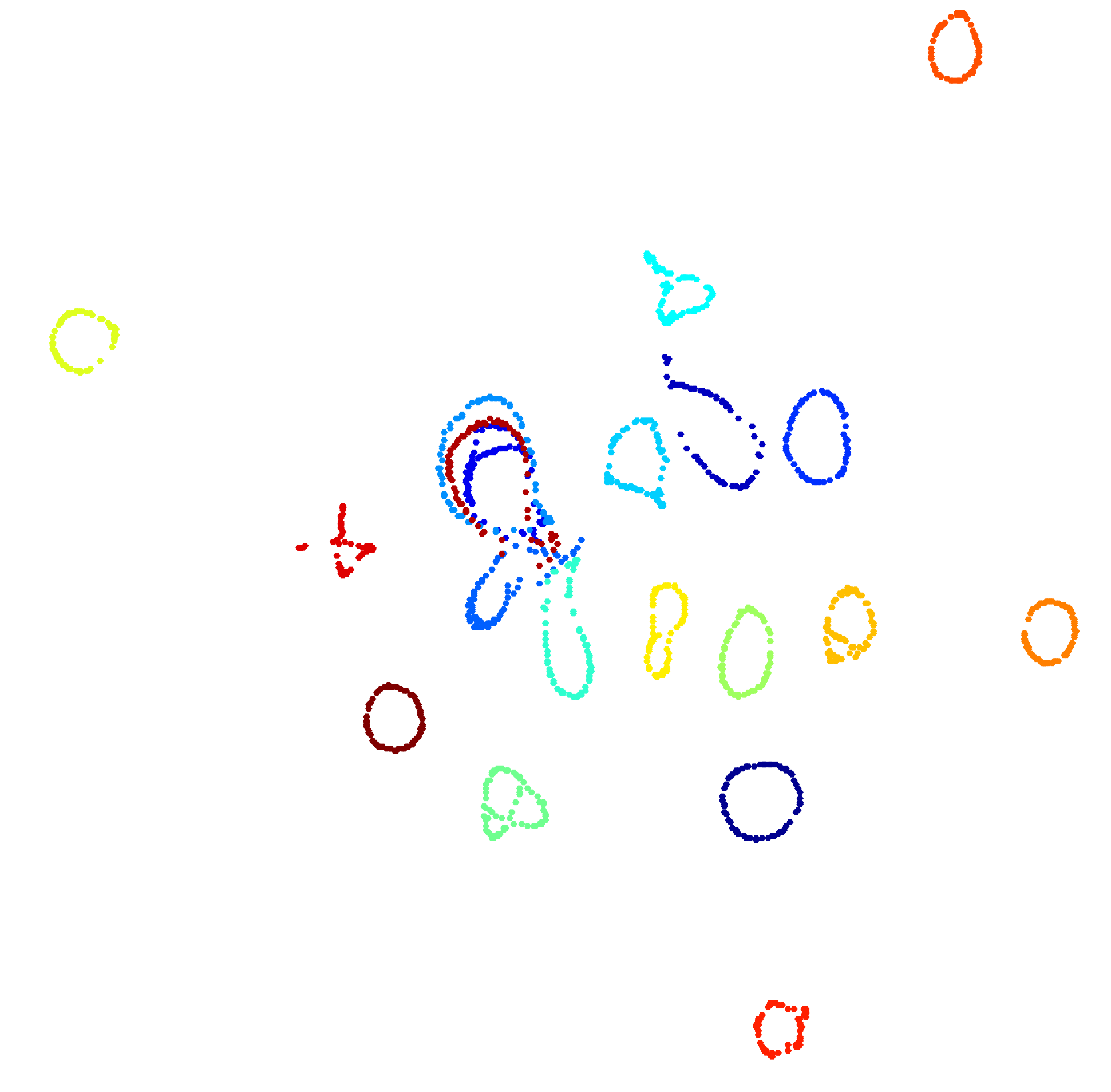}
& \includegraphics[width=0.24\textwidth]{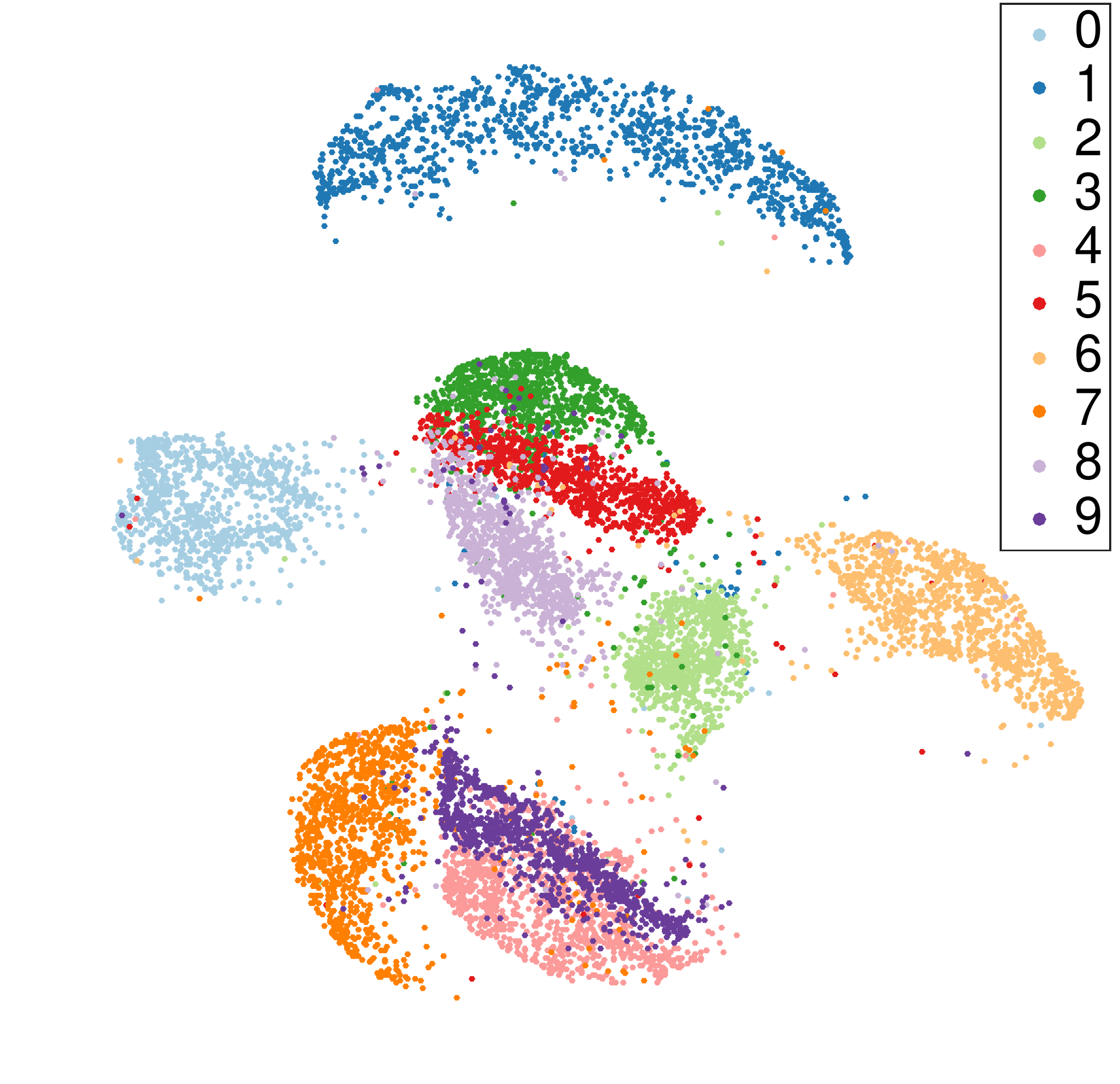}
& \includegraphics[width=0.22\textwidth]{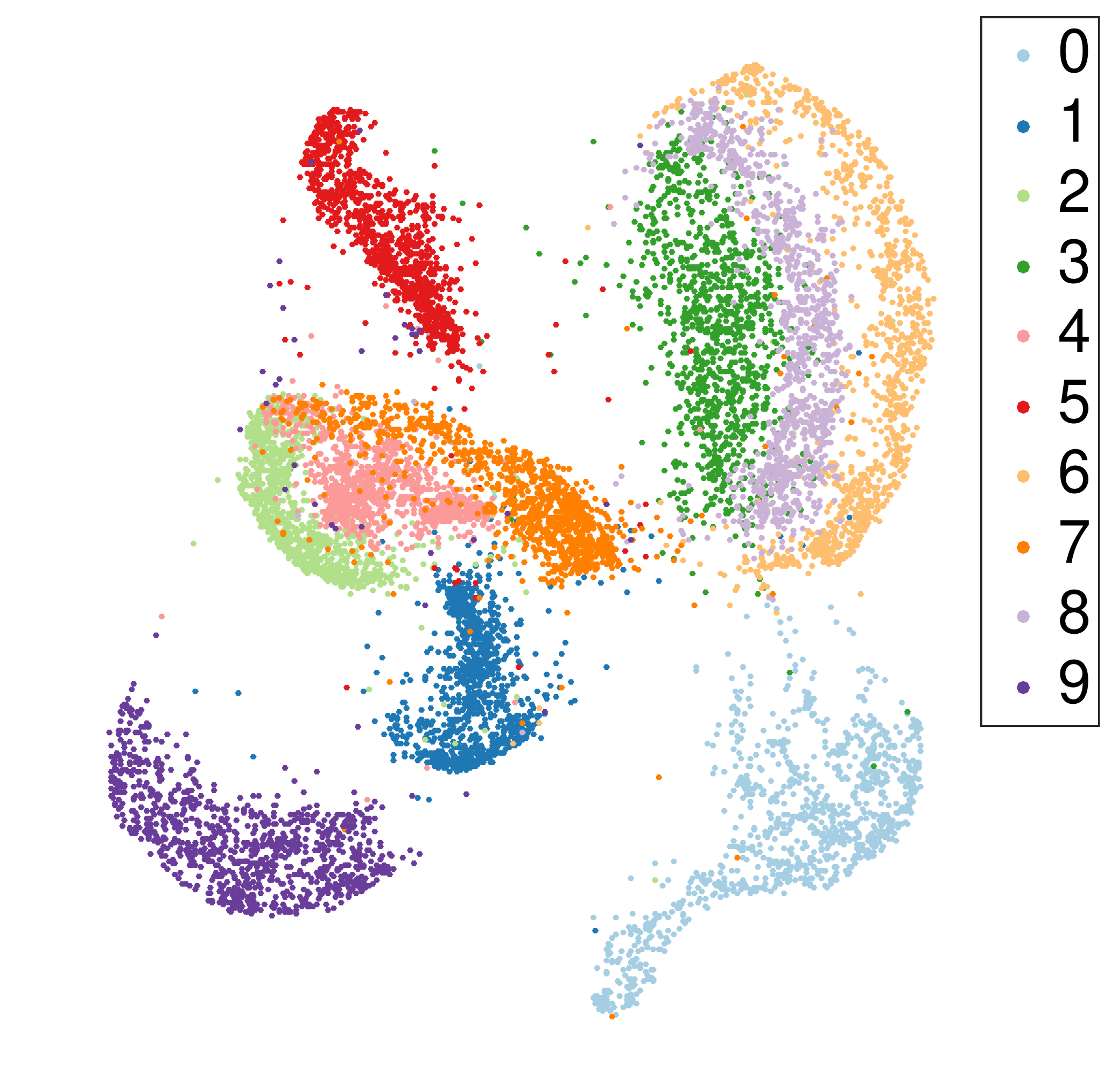}
& \includegraphics[width=0.22\textwidth]{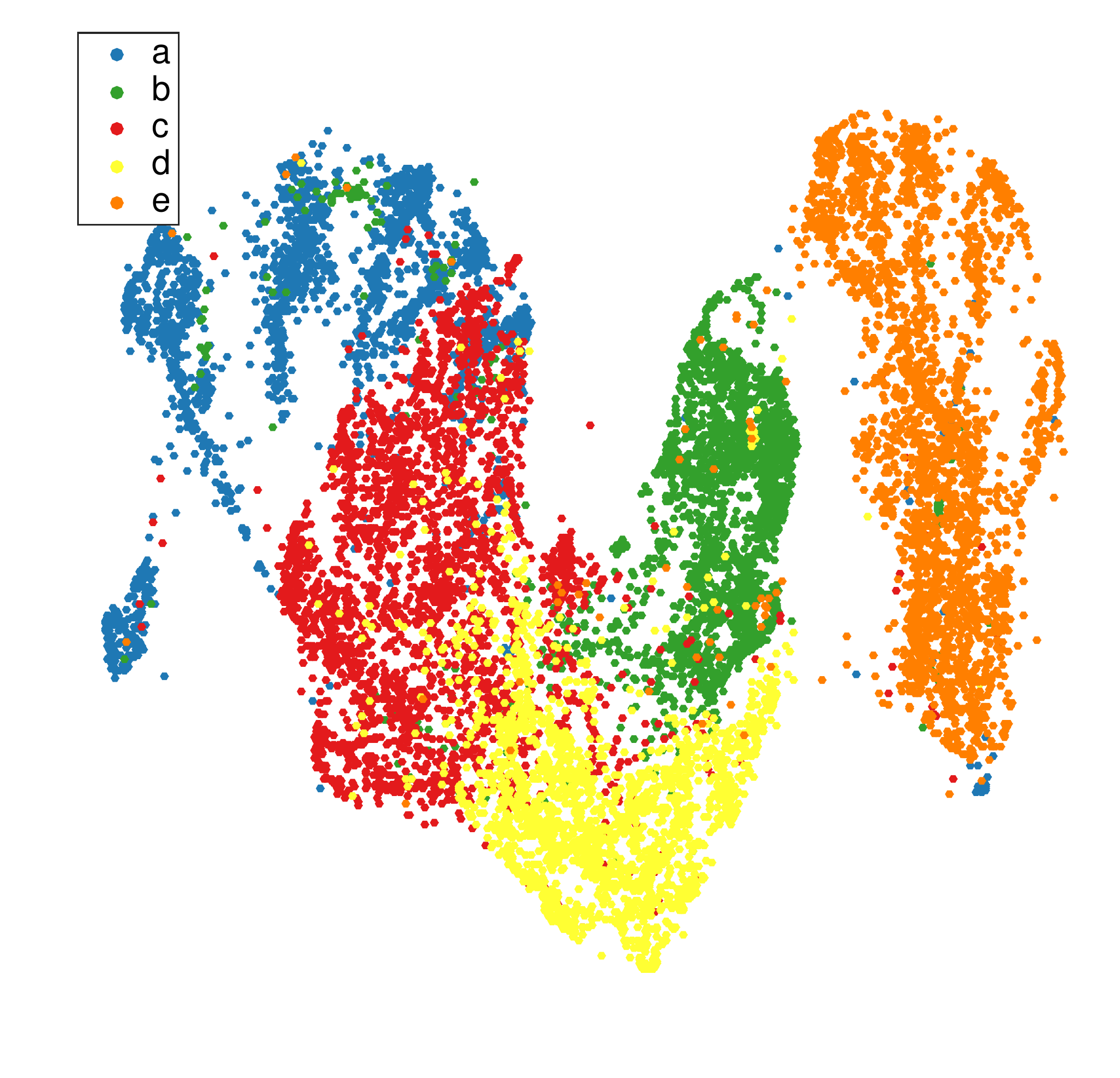}\\
  \hline
  \multicolumn{1}{c}{(e)} & \multicolumn{1}{c}{(f)} & \multicolumn{1}{c}{(g)} & \multicolumn{1}{c}{(h)}\\
\end{tabular}
\caption{Dimensionality reduction results using t-SNE (top figure) and Weighted \method\ (bottom figure) on: a) \textbf{Wine}, b) \textbf{Sphere}, c) \textbf{Swiss Roll}, d) \textbf{Faces}, e) \textbf{COIL-20}, f) \textbf{MNIST}, g) \textbf{USPS}, and h) \textbf{Letters} datasets. We use $t = t'  = 2$ in all experiments. Figures best viewed in color.}\label{fig:dr-results}
\end{center}
 \vskip -0.2in
\end{figure*}

\subsection{Generalization and Nearest-Neighbor Error}

We first evaluate the performance of different methods by means of generalization to unseen triplets as well as preserving the nearest-neighbor similarity. For this part of experiments, we consider the \textbf{MNIST Digits}~\footnote{\url{http://yann.lecun.com/exdb/mnist/}} (1000 subsamples) and \textbf{MIT Scenes}~\footnote{\url{http://people.csail.mit.edu/torralba/code/spatialenvelope/}} (800 subsamples) datasets. The synthetic triplets are generated as mentioned earlier ($100$ triplets per point). To evaluate the generalization performance, we perform a $10$-fold cross validation and report the fraction of held-out triplets that are unsatisfied as a function of number of dimension. This quantity indicates how well the method learns the underlying structure of the data. Additionally, we calculate the nearest-neighbor error as a function of number of dimensions. The nearest-neighbor error is a measure of how well the embedding captures the pairwise similarity of the objects based on relative comparisons. The results are shown in Figure~\ref{fig:gen-err}-\ref{fig:nn-err}. As can be seen, $t$-ETE performs as good as the best performing method or even better on both generalization and nearest-neighbor error. This indicates that $t$-ETE successfully captures the underlying structure of the data and scales properly with the number of dimensions. 


\subsection{Robustness to Noise}

Next, we evaluate the robustness of the different methods to  triplet noise. To evaluate the performance, we generate a different test set for both datasets with the same number of triplets as the training set. For each noise level, we randomly subsample a subset of training triplets and reverse the order of the objects. After generating the embedding, we evaluate the performance on the test set and report the fraction of the test triplets that are satisfied as well as the nearest-neighbor accuracy. The results are shown in Figure~\ref{fig:gen-err-noise}-\ref{fig:nn-err-noise}. As can be seen, the performance of all the other methods starts to drop immediately when only a small amount of noise is added to the data. On the other hand, $t$-ETE is very robust to triplet noise such that the performance is almost unaffected for up to $15\%$ of noise.  This verifies that $t$-ETE can be effectively applied to real-world datasets where a large portion of the triplets may have been corrupted by noise. 

\subsection{Visualization Results}

We provide visualization results on the \textbf{Food}~\cite{food} and \textbf{Music}~\cite{music} datasets. Figures~\ref{fig:food-tste} and~\ref{fig:food-tete} illustrate the results on the \textbf{Food} dataset using t-STE and $t$-ETE ($t = 2$), respectively. The same initialization for the data points is used for the both methods. As can be seen, no clear clusters are evident using the t-STE method. On the other hand, $t$-ETE reveals three main clusters in the data: ``Vegetables and Meals'' (top), ``Ice creams and Deserts'' (bottom left), and ``Breads and Cookies'' (bottom right).

The visualization of the \textbf{Music} dataset using the $t$-ETE method ($t = 2$) is shown in Figure~\ref{fig:music}. The result can be compared with the one using the t-STE method\footnote{Available on \url{homepage.tudelft.nl/19j49/ste}}. The distribution of the artists and the neighborhood structure are similar for both methods, but more meaningful in some regions using the $t$-ETE method. This can be due to the noise in the triplets that have been collected via human evaluators. Additionally, $t$-ETE results in $0.52$ nearest-neighbor error on the data points compared to $0.63$ error using t-STE.

\subsection{Dimensionality Reduction Results}

%

We apply the weighted triplet embedding method to find a $2$-dimensional visualization of the following datasets: $1$) \textbf{Wine}\footnote{UCI repository.}, $2$) \textbf{Sphere} ($\num{1000}$ uniform samples from a surface of a three-dimensional sphere\footnote{\url{research.cs.aalto.fi/pml/software/dredviz/}}), $3$) \textbf{Swiss Roll} ($\num{3000}$ sub-samples\footnote{\label{fn:roweis}\url{web.mit.edu/cocosci/isomap/datasets.html}}), $4$) \textbf{Faces} ($400$ synthetic faces with different pose and lighting$\text{}^\text{\ref{fn:roweis}}$), $5$) \textbf{COIL-20}\footnote{\url{www1.cs.columbia.edu/CAVE/software/softlib/coil-20.php}}, $6$) \textbf{MNIST} ($\num{10000}$ sub-samples), and $7$) \textbf{USPS} ($\num{11000}$ images of handwritten digits\footnote{\url{www.cs.nyu.edu/~roweis/data.html}}). We compare our results with those obtained using the t-SNE method. In all experiments, we use $m = 20$ for our method (for \textbf{COIL-20}, we use $m = 10$) and bias $\gamma = 0.01$. The results are shown in Figure~\ref{fig:dr-results}. 

As can be seen, our method successfully preserves the underlying structure of the data and produces high-quality embedding on all datasets, both having an underlying low-dimensional manifold (e.g., \textbf{Swiss Roll}), or clusters of points (e.g., \textbf{USPS}).  On the other hand, in most cases, t-SNE over-emphasizes the separation of the points and therefore, tears up the manifold. The same effect happens for the clusters, e.g., in the \textbf{USPS} dataset. The embedding forms multiple separated sub-clusters (for instance, the clusters of points `$3$'s, `$7$'s, and `$8$'s are divided into several smaller sub-clusters). Our objective function also enjoys better convergence properties and converges to a good solution using simple gradient descent. This eliminates the need for more complex optimization tricks such as momentum and early over-emphasis, used in t-SNE.

\section{Conclusion}
\label{sec:con}

We introduced a ranking approach for embedding a set of objects in a low-dimensional space, given a set of relative similarity constraints in the form of triplets. We showed that our method, \method, is robust to high level of noise in the triplets. We generalized our method to a weighted version to incorporate the importance of each triplet. We applied our weighted triplet embedding method to develop a new dimensionality reduction technique, which outperforms the commonly used t-SNE method in many cases while having a lower complexity and better convergence behavior.

\bibliographystyle{ACM-Reference-Format}
\bibliography{refs} 

\end{document}